\documentclass[acmtog,authorversion]{acmart}

\usepackage{booktabs} 

\usepackage{color, colortbl}
\definecolor{hicol}{rgb}{0.88,1,1}

\usepackage{multicol}
\usepackage{multirow}

\usepackage{subcaption}
\usepackage{tabularx}

\usepackage{epsfig}
\usepackage{graphicx}
\usepackage{caption}
\usepackage{subcaption}
\usepackage{stackengine}
\usepackage{tabularx}
\usepackage{rotating}
\usepackage{arydshln}

\newcommand{\etal}{et al.}

\usepackage[ruled]{algorithm2e} 

\SetAlFnt{\small}
\SetAlCapFnt{\small}
\SetAlCapNameFnt{\small}
\SetAlCapHSkip{0pt}

\begin{document}
\title[Local Light Field Fusion]{Local Light Field Fusion: \protect\\Practical View Synthesis with Prescriptive Sampling Guidelines}
 
\author{Ben Mildenhall}
\authornote{Denotes equal contribution}
\affiliation{
  \institution{University of California, Berkeley}}

\author{Pratul P. Srinivasan}
\authornotemark[1]
\affiliation{
  \institution{University of California, Berkeley}}

\author{Rodrigo Ortiz-Cayon}
\affiliation{
  \institution{Fyusion Inc.}}

\author{Nima Khademi Kalantari}
\affiliation{
  \institution{Texas A\&M University}}

\author{Ravi Ramamoorthi}
\affiliation{
  \institution{University of California, San Diego}}

\author{Ren Ng}
\affiliation{
  \institution{University of California, Berkeley}}

\author{Abhishek Kar}
\affiliation{
  \institution{Fyusion Inc.}}
  
\renewcommand{\shortauthors}{B. Mildenhall, P. P. Srinivasan, R. Ortiz-Cayon, N. Khademi Kalantari, R. Ramamoorthi, R. Ng, and A. Kar}

\begin{abstract}
We present a practical and robust deep learning solution for capturing and rendering novel views of complex real world scenes for virtual exploration. Previous approaches either require intractably dense view sampling or provide little to no guidance for how users should sample views of a scene to reliably render high-quality novel views. Instead, we propose an algorithm for view synthesis from an irregular grid of sampled views that first expands each sampled view into a local light field via a multiplane image (MPI) scene representation, then renders novel views by blending adjacent local light fields. We extend traditional plenoptic sampling theory to derive a bound that specifies precisely how densely users should sample views of a given scene when using our algorithm. In practice, we apply this bound to capture and render views of real world scenes that achieve the perceptual quality of Nyquist rate view sampling while using up to $4000\times$ fewer views. We demonstrate our approach's practicality with an augmented reality smartphone app that guides users to capture input images of a scene and viewers that enable realtime virtual exploration on desktop and mobile platforms.
\end{abstract}

%
%


\setcopyright{acmlicensed}
\acmJournal{TOG}
\acmYear{2019}\acmVolume{38}\acmNumber{4}\acmArticle{29}\acmMonth{7} \acmDOI{10.1145/3306346.3322980}

%
%


\begin{teaserfigure}
  \centering
  \includegraphics[width=\textwidth]{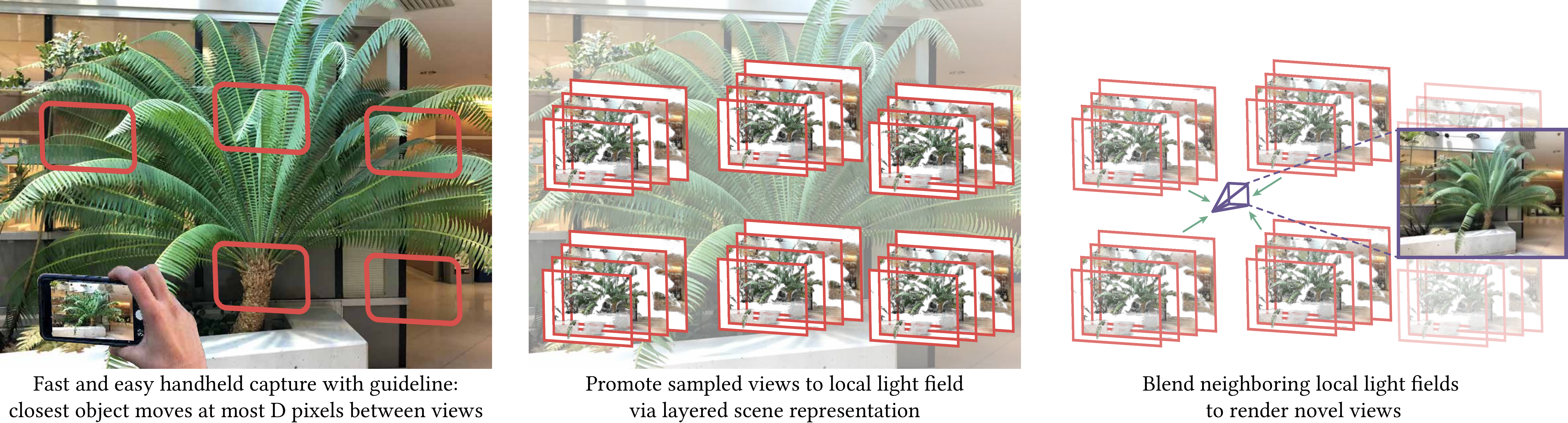}
  \caption{
  We present a simple and reliable method for view synthesis from a set of input images captured by a handheld camera in an irregular grid pattern. We theoretically and empirically demonstrate that our method enjoys a prescriptive sampling rate that requires ~$4000\times$ fewer input views than Nyquist for high-fidelity view synthesis of natural scenes. Specifically, we show that this rate can be interpreted as a requirement on the pixel-space disparity of the closest object to the camera between captured views (Section~\ref{sec:sampling}). After capture, we expand all sampled views into layered representations that can render high-quality local light fields. We then blend together renderings from adjacent local light fields to synthesize dense paths of new views (Section~\ref{sec:pipeline}). Our rendering consists of simple and fast computations (homography warping and alpha compositing) that can generate new views in real-time.
  } 
  \label{fig:teaser}
\end{teaserfigure}

\maketitle

\section{Introduction}

The most compelling virtual experiences completely immerse the viewer in a scene, and a hallmark of such experiences is the ability to view the scene from a close interactive distance. This is currently possible with synthetically rendered scenes, but this level of intimacy has been very difficult to achieve for virtual experiences of real world scenes.

Ideally, we could simply sample the scene's light field and interpolate the relevant captured images to render new views. Such light field sampling strategies are particularly appealing because they pose the problem of image-based rendering (IBR) in a signal processing framework where we can directly reason about the density and pattern of sampled views required for any given scene. However, Nyquist rate view sampling is intractable for scenes with content at interactive distances, as the required view sampling rate increases linearly with the reciprocal of the closest scene depth. For example, for a scene with a subject at a depth of 0.5 meters captured by a mobile phone camera with a $64^{\circ}$ field of view and rendered at 1 megapixel resolution, the required sampling rate is an intractable 2.5 million images per square meter. Since it is not feasible to capture all the required images, the IBR community has moved towards view synthesis algorithms that leverage geometry estimation to predict the missing views. 

State-of-the-art algorithms pose the view synthesis problem as the prediction of novel views from an unstructured set or arbitrarily sparse grid of input camera views. While the generality of this problem statement is appealing, abandoning a plenoptic sampling framework sacrifices the crucial ability to rigorously reason about the view sampling requirements of these methods and predict how their performance will be affected by the input view sampling pattern. When faced with a new scene, users of these methods are limited to trial-and-error to figure out whether a set of sampled views will produce acceptable results for a virtual experience.

Instead, we propose a view synthesis approach that is grounded within a plenoptic sampling framework and can precisely prescribe how densely a user must capture a given scene for reliable rendering performance. Our method is conceptually simple and consists of two main stages. We first use a deep network to promote each source view to a layered representation of the scene that can render a limited range of views, advancing recent work on the multiplane image (MPI) representation~\cite{zhou18}. We then synthesize novel views by blending renderings from adjacent layered representations.

Our theoretical analysis shows that the number of input views required by our method decreases quadratically with the number of planes we predict for each layered scene representation, up to limits set by the camera field of view. We empirically validate our analysis and apply it in practice to render novel views with the same perceptual quality as Nyquist view sampling while using up to $64^2\approx 4000 \times$ fewer images.

It is impossible to break the Nyquist limit with full generality, but we show that it is possible to achieve Nyquist level performance with greatly reduced view sampling by specializing to the subset of natural scenes. 
This capability is primarily due to our deep learning pipeline, which is trained on renderings of natural scenes to estimate high quality layered scene representations that produce locally consistent light fields.

In summary, our key contributions are:
\begin{enumerate}
\vspace{-1mm}
\item An extension of plenoptic sampling theory that directly specifies how users should sample input images for reliable high quality view synthesis with our method.
\item A practical and robust solution for capturing and rendering complex real world scenes for virtual exploration.
\item A demonstration that carefully crafted deep learning pipelines using local layered scene representations achieve state-of-the-art view synthesis results.
\end{enumerate}

We extensively validate our derived prescriptive view sampling requirements and demonstrate that our algorithm quantitatively outperforms traditional light field reconstruction methods as well as state-of-the-art view interpolation algorithms across a range of sub-Nyquist view sampling rates. We highlight the practicality of our method by developing an augmented reality app that implements our derived sampling guidelines to help users capture input images that produce reliably high-quality renderings with our algorithm. Additionally, we develop mobile and desktop viewer apps that render novel views from our predicted layered representations in real-time. Finally, we qualitatively demonstrate that our algorithm reliably produces state-of-the-art results across a diverse set of complex real-world scenes.

\section{Related Work}

Image-based rendering (IBR) is the fundamental computer graphics problem of rendering novel views of objects and scenes from sampled views. We find that it is useful to categorize IBR algorithms by the extent to which they use explicit scene geometry, as done by Shum and Kang~\shortcite{shum00}.

\subsection{Plenoptic Sampling and Reconstruction}

Light field rendering~\cite{levoy96} eschews any geometric reasoning and simply samples images on a regular grid so that new views can be rendered as slices of the sampled light field. Lumigraph rendering~\cite{gortler96} showed that using approximate scene geometry can ameliorate artifacts due to undersampled or irregularly sampled views.

The plenoptic sampling framework~\cite{chai00} analyzes light field rendering using signal processing techniques and shows that the Nyquist view sampling rate for light fields depends on the minimum and maximum scene depths. Furthermore, they discuss how the Nyquist view sampling rate can be lowered with more knowledge of scene geometry. Zhang and Chen~\shortcite{zhang2003} extend this analysis to show how non-Lambertian and occlusion effects increase the spectral support of a light field, and also propose more general view sampling lattice patterns. 

Rendering algorithms based on plenoptic sampling enjoy the significant benefit of prescriptive sampling; given a new scene, it is easy to compute the required view sampling density to enable high-quality renderings. Many modern light field acquisition systems have been designed based on these principles, including large-scale camera systems~\cite{wilburn05,overbeck2018} and a mobile phone app~\cite{davis2012}.

We posit that prescriptive sampling is necessary for practical and useful IBR algorithms, and we extend prior theory on plenoptic sampling to show that our deep-learning-based view synthesis strategy can significantly decrease the dense sampling requirements of traditional light field rendering. Our novel view synthesis pipeline can also be used in future light field acquisition hardware systems to reduce the number of required cameras.

\subsection{Geometry-Based View Synthesis}

Many IBR algorithms attempt to leverage explicit scene geometry to synthesize new views from arbitrary unstructured sets of input views. These approaches can be meaningfully categorized as either using global or local geometry.

Techniques that use global geometry generally compute a single global mesh from a set of unstructured input images. Simply texture mapping this global mesh can be effective for constrained situations such as panoramic viewing with mostly rotational and little translational viewer movement~\cite{hedman17,hedman18}, but this strategy can only simulate Lambertian materials. Surface light fields~\cite{wood00} are able to render convincing view-dependent effects, but they require accurate geometry from dense range scans and hundreds of captured images to sample the outgoing radiance at points on an object's surface.

Many free-viewpoint IBR algorithms are based upon a strategy of locally texture mapping a global mesh. The influential view-dependent texture mapping algorithm~\cite{debevec96} proposed an approach to render novel views by blending nearby captured views that have been reprojected using a global mesh. Work on Unstructured Lumigraph Rendering~\cite{buehler01} focused on computing per-pixel blending weights for reprojected images and proposed a heuristic algorithm that satisfied key properties for high-quality rendering. Unfortunately, it is very difficult to estimate high-quality meshes whose geometric boundaries align well with edges in images, and IBR algorithms based on global geometry typically suffer from significant artifacts. State-of-the-art algorithms~\cite{hedman16,hedman182} attempt to remedy this shortcoming with complicated pipelines that involve both global mesh and local depth map estimation. However, it is difficult to precisely define view sampling requirements for robust mesh estimation, and the mesh estimation procedure typically takes multiple hours, making this strategy impractical for casual content capture scenarios.

IBR algorithms that use local geometry~\cite{chaurasia2013,chen93,kopf13,mcmillan95,ortizcayon15} avoid difficult and expensive global mesh estimation. Instead, they typically compute detailed local geometry for each input image and render novel views by reprojecting and blending nearby input images. This strategy has also been extended to simulate non-Lambertian reflectance by using a second depth layer~\cite{sinha12}. The state-of-the-art Soft3D algorithm~\cite{penner17} blends between reprojected local layered representations to render novel views, which is conceptually similar to our strategy. However, Soft3D computes each local layered representation by aggregating heuristic measures of depth uncertainty over a large neighborhood of views. 
We instead train a deep learning pipeline end-to-end to optimize novel view quality by predicting each of our local layered representations from a much smaller neighborhood of views. 
Furthermore, we directly pose our algorithm within a plenoptic sampling framework, and our analysis directly applies to the Soft3D algorithm as well. We demonstrate that the high quality of our deep learning predicted local scene representations allows us to synthesize superior renderings without requiring the aggregation of geometry estimates over large view neighborhoods, as done in Soft3D. This is especially advantageous for rendering non-Lambertian effects because the apparent depth of specularities generally varies with the observation viewpoint, so smoothing the estimated geometry over large viewpoint neighborhoods prevents accurate rendering of these effects. 

Other IBR algorithms~\cite{anderson16} have attempted to be more robust to incorrect camera poses or scene motion by interpolating views using more general 2D optical flow instead of 1D depth. Local pixel shifts are also encoded in the phase information, and algorithms have exploited this to extrapolate views from micro-baseline stereo pairs~\cite{didyk13,kellnhofer17,zhang15} without explicit flow computation. However, these methods require extremely close input views and are not suited for large baseline view interpolation.

\subsection{Deep Learning for View Synthesis}

Other recent methods have trained deep learning pipelines end-to-end for view synthesis. This includes recent angular superresolution methods~\cite{wu17,yeung18} that interpolate dense views within a light field camera's aperture but cannot handle sparser input view sampling since they do not model scene geometry.
The DeepStereo algorithm~\cite{flynn16}, deep learning based light field camera view interpolation~\cite{kalantari16}, and single view local light field synthesis~\cite{srinivasan17} each use a deep network to predict depth separately for every novel view. However, predicting local geometry separately for each view results in inconsistent renderings across smoothly-varying viewpoints.

Finally, Zhou \etal~\shortcite{zhou18} introduce a deep learning pipeline to predict an MPI from a narrow baseline stereo pair for the task of stereo magnification. As opposed to previous deep learning strategies for view synthesis, this approach enforces consistency by using the same predicted scene representation to render all novel views. We adopt MPIs as our local light field representation and introduce specific technical improvements to enable larger-baseline view interpolation from many input views, in contrast to local view extrapolation from a stereo pair using a single MPI. We predict multiple MPIs, one for each input view, and train our system end-to-end through a blending procedure to optimize the resulting MPIs to be used in concert for rendering output views. We propose a 3D convolutional neural network (CNN) architecture that dynamically adjusts the number of depth planes based on the input view sampling rate, rather than a 2D CNN with a fixed number of output planes. Additionally, we show that state-of-the-art performance requires only an easily-generated synthetic dataset and a small real fine-tuning dataset, rather than a large real dataset. This allows us to generate training data captured on 2D irregular grids similar to handheld view sampling patterns, while the YouTube dataset in Zhou \etal~\shortcite{zhou18} is restricted to 1D camera paths.

\section{Theoretical Sampling Analysis}
\label{sec:sampling}

The overall strategy of our method is to use a deep learning pipeline to promote each sampled view to a layered scene representation with $D$ depth layers, and render novel views by blending between renderings from neighboring scene representations. In this section, we show that the full set of scene representations predicted by our deep network can be interpreted as a specific form of light field sampling. We extend prior work on plenoptic sampling to show that our strategy can theoretically reduce the number of required sampled views by a factor of $D^{2}$ compared to the number required by traditional Nyquist view sampling. Section~\ref{sec:sampling_validation} empirically shows that we are able to take advantage of this bound to reduce the number of required views by up to $64^{2}\approx4000\times$.

In the following analysis, we consider a ``flatland'' light field with a single spatial dimension $x$ and view dimension $u$ for notational clarity, but note that all findings apply to general light fields with two spatial and two view dimensions.

\subsection{Nyquist Rate View Sampling}

\begin{table}[t!]
    \centering
    \caption{Reference for symbols used in Section~\ref{sec:sampling}.}
    \vspace{-0.1in}
    \begin{tabular}{|c|l|}
    \hline Symbol   &  Definition\\ \hline
       $D$  & Number of depth planes \\ 
       $W$  & Camera image width (pixels) \\ 
       $f$ & Camera focal length (meters) \\ 
       $\Delta_x$  & Pixel size (meters) \\ 
       $\Delta_u$  & Baseline between cameras (meters) \\ 
       $K_x$  & Highest spatial frequency in sampled light field \\ 
       $B_x$  & Highest spatial frequency in continuous light field \\ 
       $z_{\min}$  & Closest scene depth (meters) \\
       $z_{\max}$  & Farthest scene depth (meters) \\
       $d_{\max}$  & Maximum disparity between views (pixels) \\
       \hline
    \end{tabular}
    \label{table:symbols}
\end{table}

\begin{figure}[t]
\captionsetup[subfigure]{font=scriptsize,labelformat=empty,aboveskip=1pt,belowskip=0pt}
  \centering
  \begin{subfigure}[t]{0.592\columnwidth}
    \centering\includegraphics[width=\textwidth]{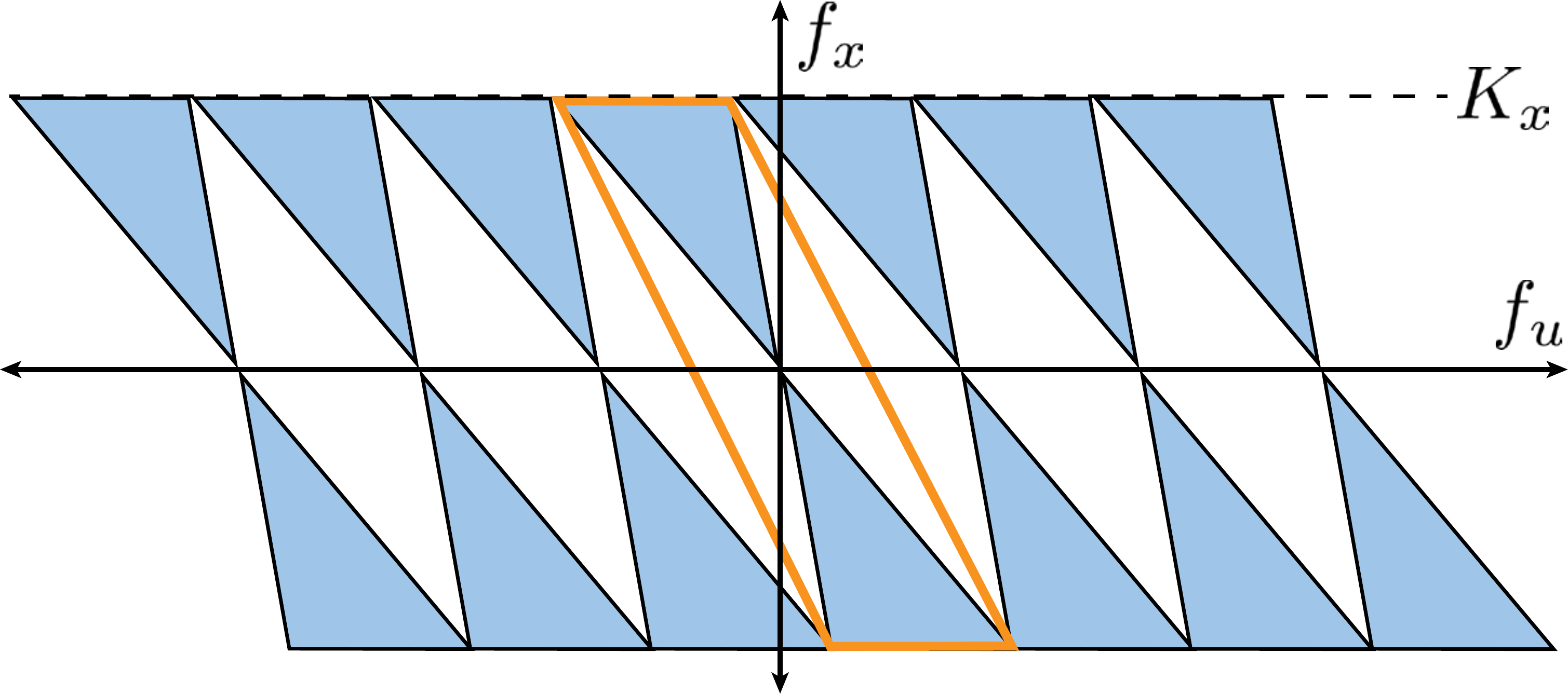}
    \caption{(a)}
  \end{subfigure}
  \begin{subfigure}[t]{0.213\columnwidth}
    \centering\includegraphics[width=\textwidth]{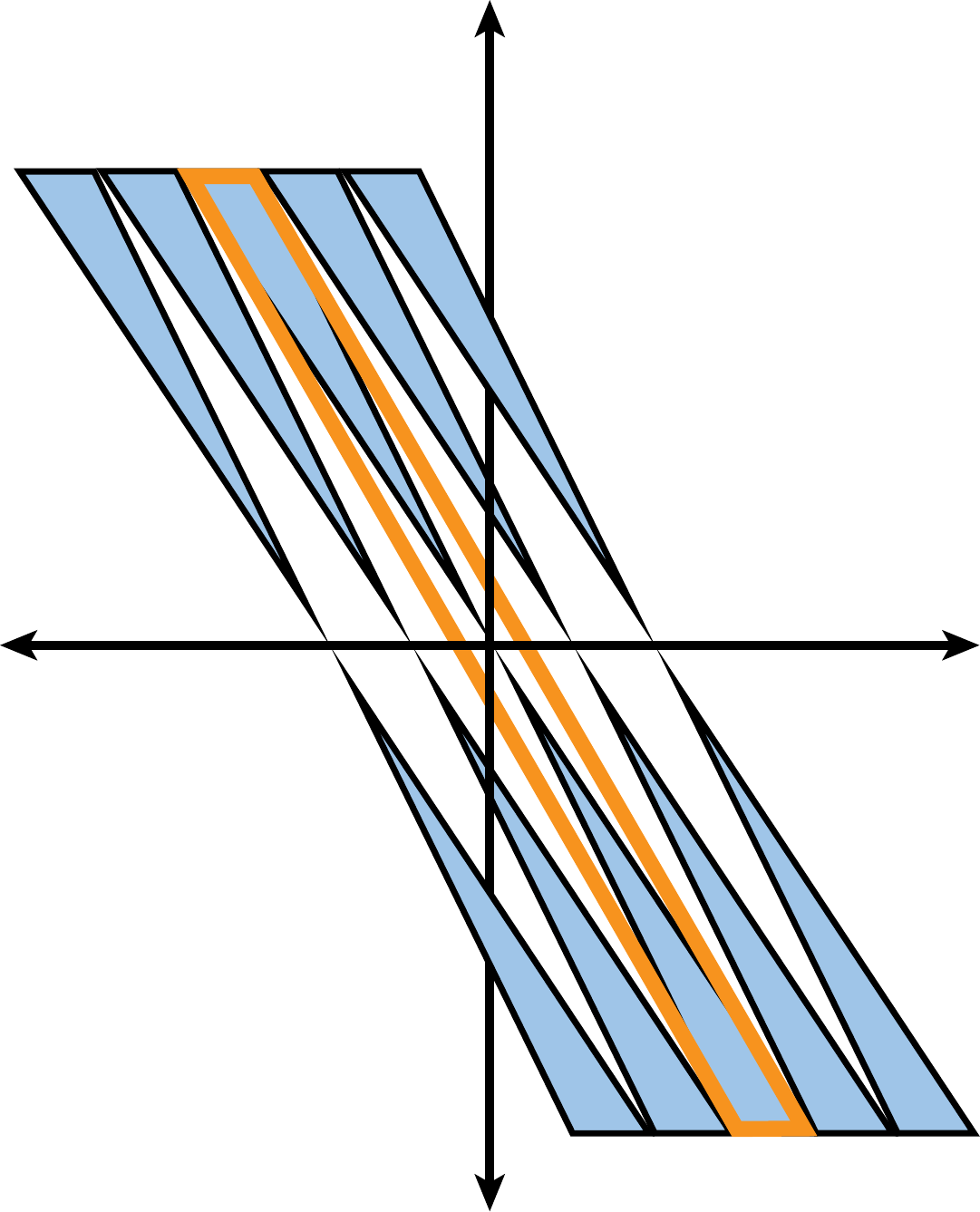}
    \caption{(b)}
  \end{subfigure}
  \begin{subfigure}[t]{0.177\columnwidth}
    \centering\includegraphics[width=\textwidth]{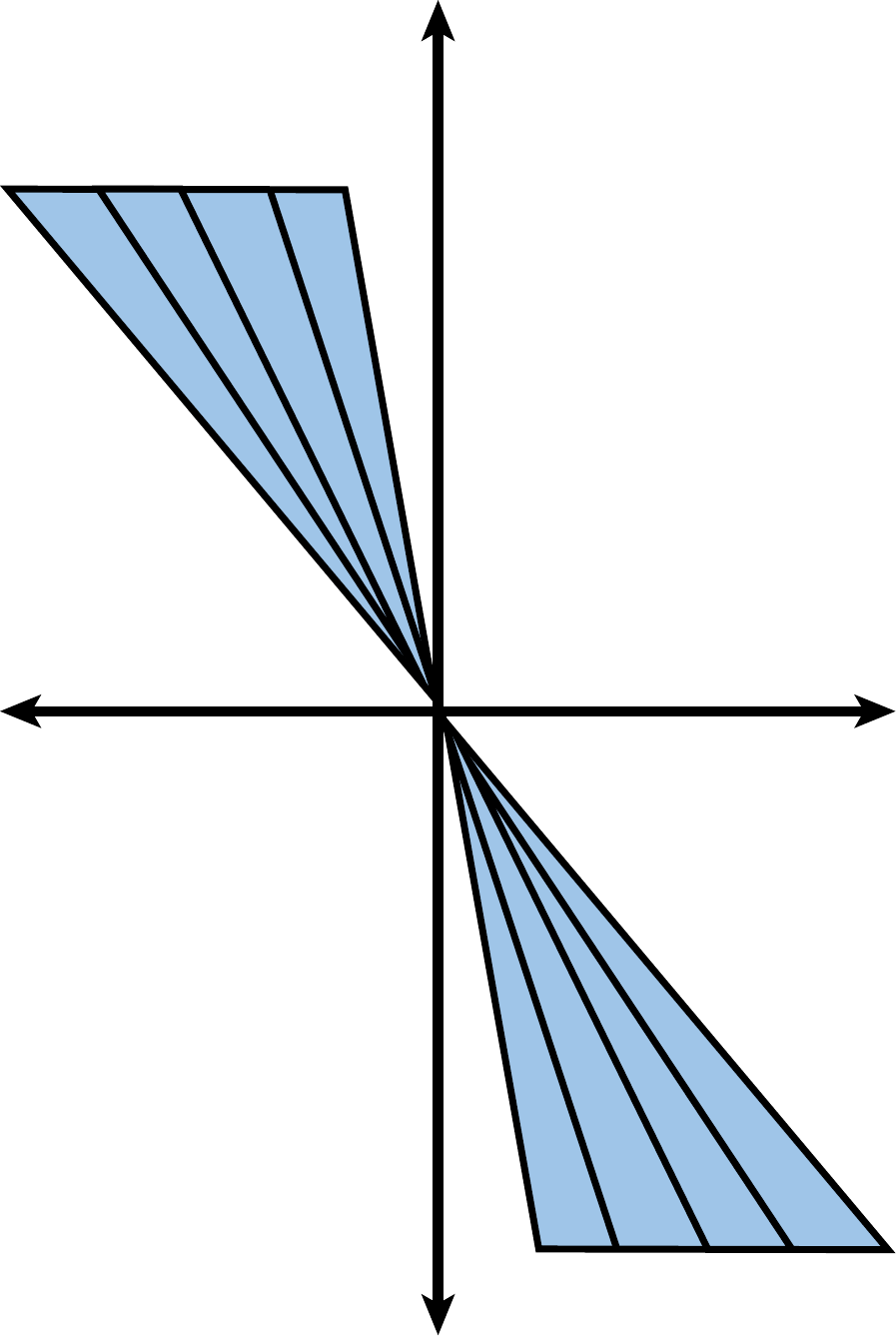}
    \caption{(c)}
  \end{subfigure}
  
\vspace{-3mm}
  \caption{Traditional plenoptic sampling without occlusions, as derived in~\cite{chai00}. (a) The Fourier support of a light field without occlusions lies within a double-wedge, shown in blue. Nyquist rate view sampling is set by the double-wedge width, which is determined by the minimum and maximum scene depths $[z_{\min},z_{\max}]$ and the maximum spatial frequency $K_x$. The ideal reconstruction filter is shown in orange. (b) Splitting the light field into $D$ non-overlapping layers with equal disparity width decreases the Nyquist rate by a factor of $D$. (c) Without occlusions, the full light field spectrum is the sum of the spectra from each layer.}
\label{fig:wedge}
 \vspace{-3mm}
\end{figure}

\begin{figure}
\captionsetup[subfigure]{font=scriptsize,labelformat=empty,aboveskip=1pt,belowskip=0pt}
  \centering
  \begin{subfigure}[t]{0.552\columnwidth}
    \centering\includegraphics[width=\textwidth]{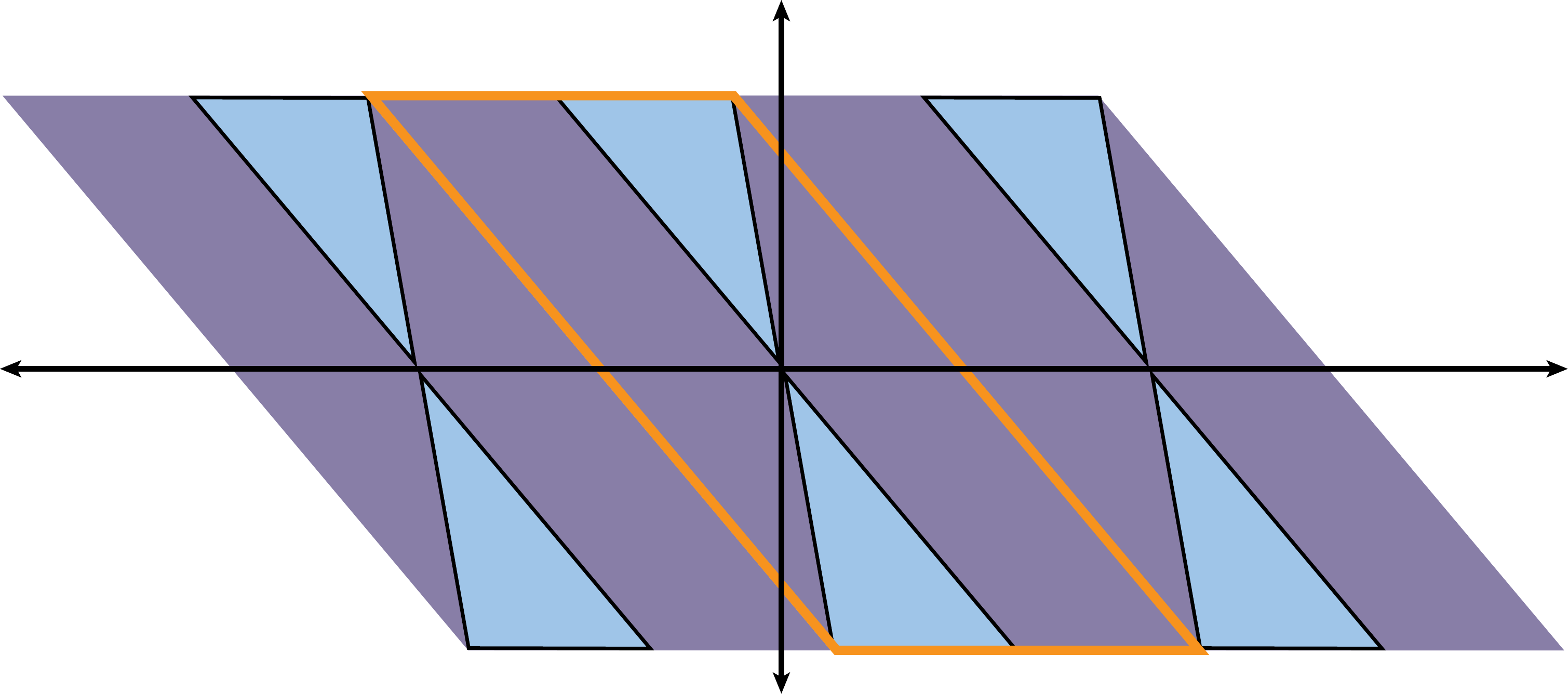}
    \caption{(a)}
  \end{subfigure}
  \begin{subfigure}[t]{0.232\columnwidth}
    \centering\includegraphics[width=\textwidth]{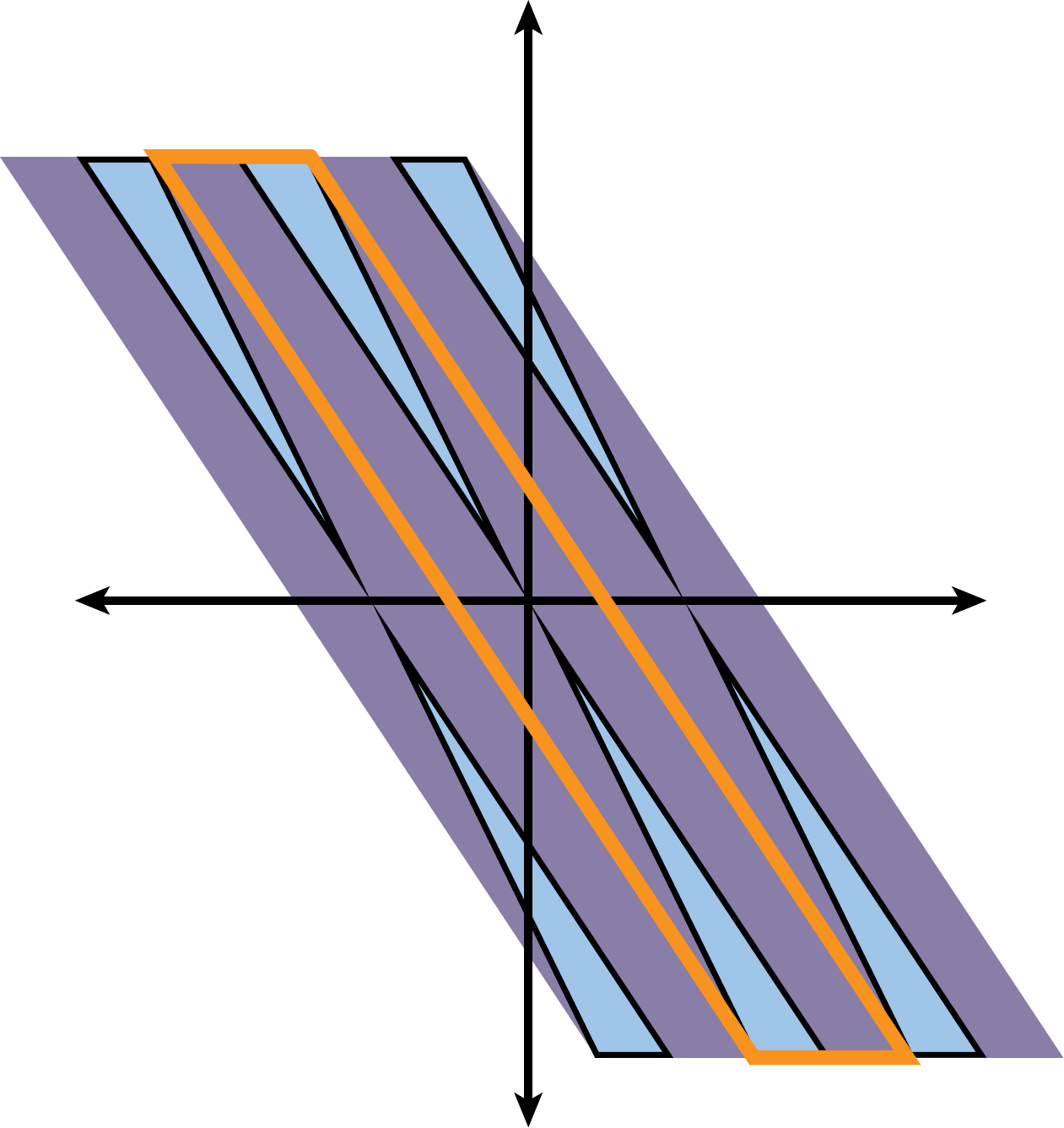}
    \caption{(b)}
  \end{subfigure}
  \begin{subfigure}[t]{0.198\columnwidth}
    \centering\includegraphics[width=\textwidth]{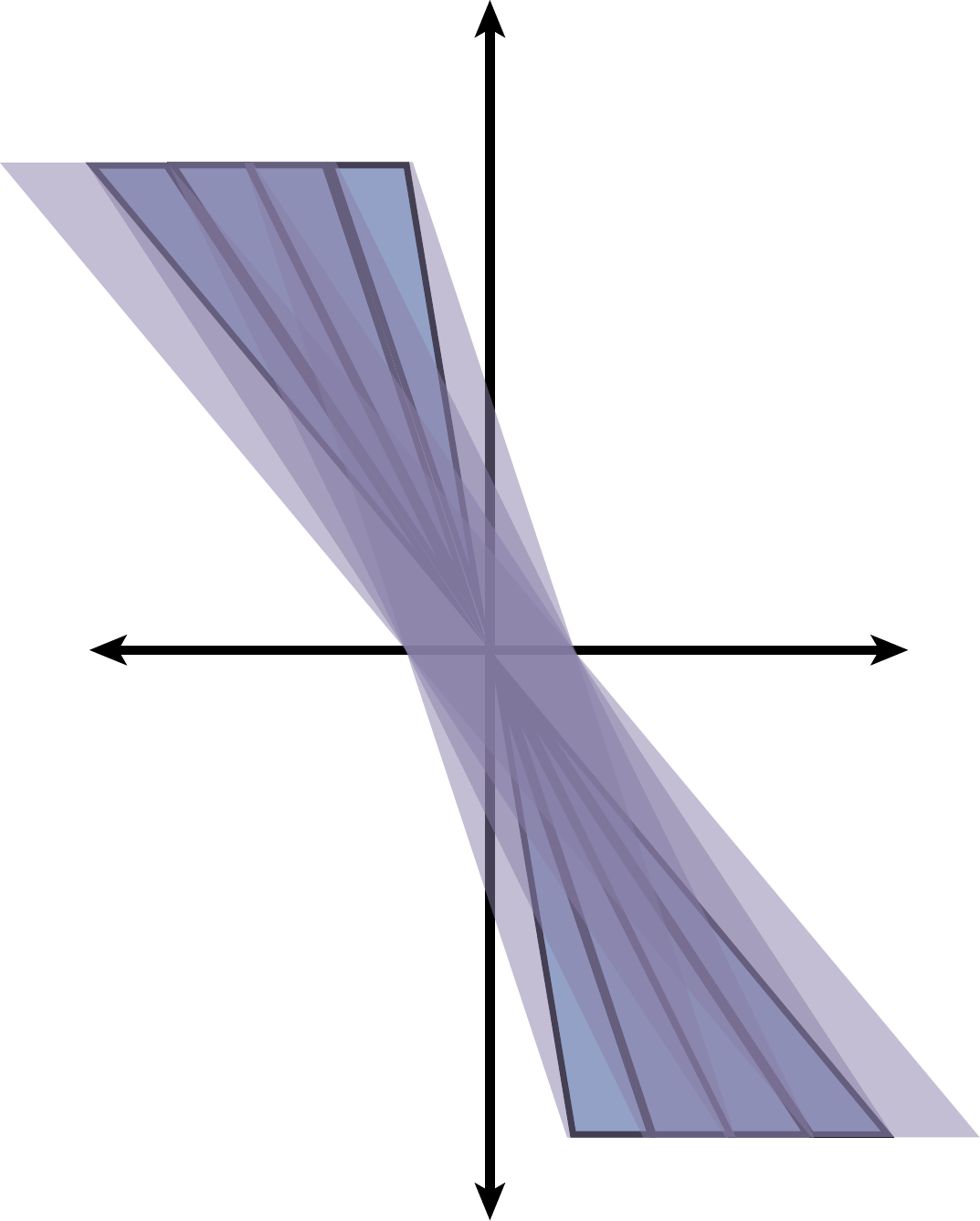}
    \caption{(c)}
  \end{subfigure}
  
\vspace{-3mm}
  \caption{We extend traditional plenoptic sampling to consider occlusions when reconstructing a continuous light field from MPIs. (a) Considering occlusions expands the Fourier support to a parallelogram (the Fourier support without occlusions is shown in blue and occlusions expand the Fourier support to additionally include the purple region) and doubles the Nyquist view sampling rate. (b) As in the no-occlusions case, separately reconstructing the light field for $D$ layers decreases the Nyquist rate by a factor of $D$. (c) With occlusions, the full light field spectrum cannot be reconstructed by summing the individual layer spectra because the union of their supports is smaller than the support of the full light field spectrum (a). Instead, we compute the full light field by alpha compositing the individual light field layers from back to front in the primal domain.}
\label{fig:wedge_occ}

\end{figure}

Initial work on plenoptic sampling~\cite{chai00} derived that the Fourier support of a light field, ignoring occlusion and non-Lambertian effects, lies within a double-wedge shape whose bounds are set by the minimum and maximum scene depths $z_{\min}$ and $z_{\max}$, as visualized in Figure~\ref{fig:wedge}. Zhang and Chen~\shortcite{zhang2003} showed that occlusions expand the light field's Fourier support because an occluder convolves the spectrum of the light field due to farther scene content with a kernel that lies on the line corresponding to the occluder's depth. The light field's Fourier support considering occlusions is limited by the effect of the closest occluder convolving the line corresponding to the furthest scene content, resulting in the parallelogram shape illustrated in Figure~\ref{fig:wedge_occ}a, which can only be packed half as densely as the double-wedge. The required maximum camera sampling interval $\Delta_{u}$ for a light field with occlusions is:
\begin{equation}
    \Delta_{u} \leq \frac{1}{2K_{x}f\left(1/z_{\min}-1/z_{\max}\right)}.
\end{equation}
$K_{x}$ is the highest spatial frequency represented in the sampled light field, determined by the highest spatial frequency in the continuous light field $B_{x}$ and the camera spatial resolution $\Delta_{x}$:
\begin{align}
    K_{x}=\min \left(B_{x},\frac{1}{2\Delta_{x}}\right).
\end{align}

\subsection{MPI Scene Representation and Rendering}

\begin{figure}
\centering
    \includegraphics[width=\columnwidth]{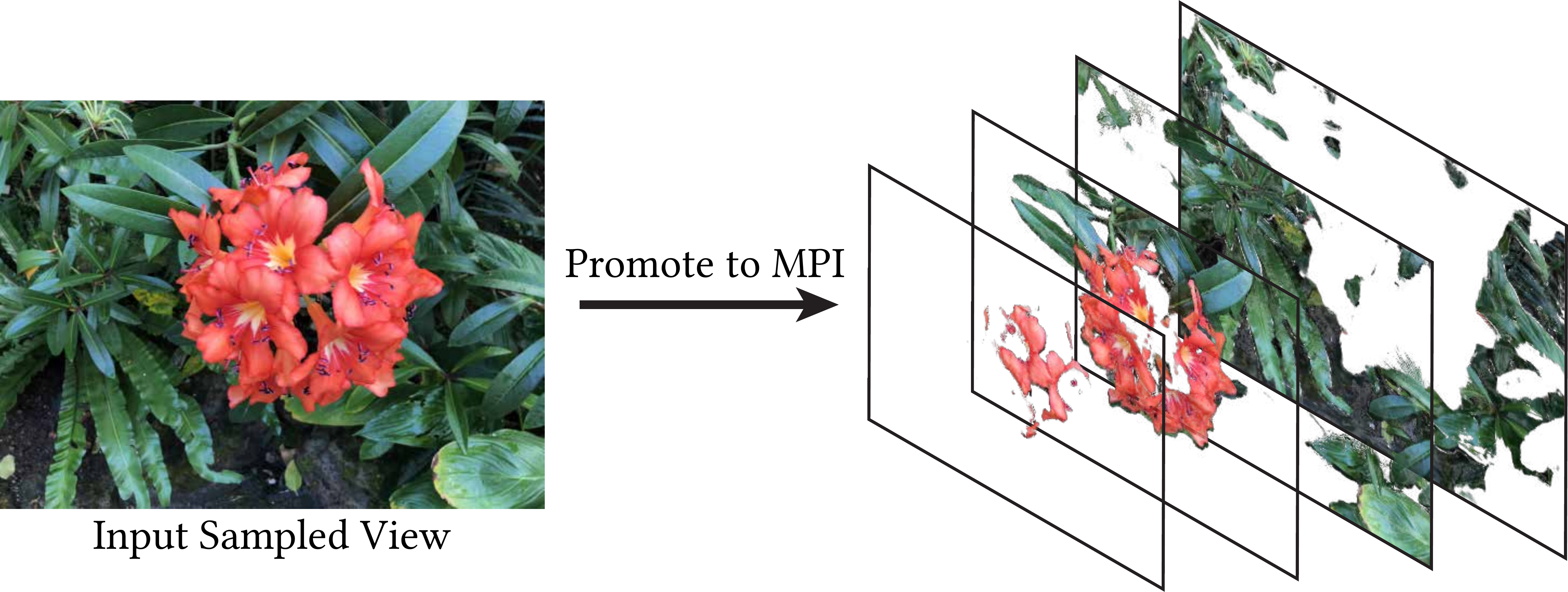}
    \caption{We promote each input view sample to an MPI scene representation~\cite{zhou18}, consisting of $D$ RGB$\alpha$ planes at regularly sampled disparities within the input view's camera frustum. Each MPI can render continuously-valued novel views within a local neighborhood by alpha compositing color along rays into the novel view's camera.}
    \label{fig:mpi_viz}
\end{figure}

The MPI scene representation~\cite{zhou18} consists of a set of fronto-parallel RGB$\alpha$ planes, evenly sampled in disparity within a reference camera's view frustum (see Figure~\ref{fig:mpi_viz}). We can render novel views from an MPI at continuously-valued camera poses within a local neighborhood by alpha compositing the color along rays into the novel view camera using the ``over'' operator~\cite{porter84}. This rendering procedure is equivalent to reprojecting each MPI plane onto the sensor plane of the novel view camera and alpha compositing the MPI planes from back to front, as observed in early work on volume rendering~\cite{lacroute94}. An MPI can be considered as an encoding of a local light field, similar to layered light field displays~\cite{wetzstein11,wetzstein12}.

\subsection{View Sampling Rate Reduction}

Plenoptic sampling theory~\cite{chai00} additionally shows that decomposing a scene into $D$ depth ranges and separately sampling the light field within each range allows the camera sampling interval to be increased by a factor of $D$. This is because the spectrum of the light field emitted by scene content within each depth range lies within a tighter double-wedge that can be packed $D$ times more tightly than the full scene's double-wedge spectrum. Therefore, a tighter reconstruction filter with a different shear can be used for each depth range, as illustrated in Figure~\ref{fig:wedge}b. The reconstructed light field, ignoring occlusion effects, is simply the sum of the reconstructions of all layers, as shown in Figure~\ref{fig:wedge}c. 

However, it is not straightforward to extend this analysis to handle occlusions, because the union of the Fourier spectra for all depth ranges has a smaller support than the original light field with occlusions, as visualized in Figure~\ref{fig:wedge_occ}c. Instead, we observe that reconstructing a full scene light field from these depth range light fields while respecting occlusions would be much easier given corresponding per-view opacities, or shield fields~\cite{lanman08}, for each layer. We could then easily alpha composite the depth range light fields from back to front to compute the full scene light field.

Each alpha compositing step increases the Fourier support by convolving the previously-accumulated light field's spectrum with the spectrum of the occluding depth layer. As is well known in signal processing, the convolution of two spectra has a Fourier bandwidth equal to the sum of the original spectra's bandwidths. Figure~\ref{fig:wedge_occ}b illustrates that the width of the Fourier support parallelogram for each depth range light field, considering occlusions, is:
\begin{equation}
    2K_{x}f\left(1/z_{\min}-1/z_{\max}\right)/D,
\end{equation}
so the resulting reconstructed light field of the full scene will enjoy the full Fourier support width.

We apply this analysis to our algorithm by interpreting the predicted MPI layers at each camera sampling location as view samples of scene content within non-overlapping depth ranges, and noting that applying the optimal reconstruction filter~\cite{chai00} for each depth range is equivalent to reprojecting and then blending pre-multiplied RGB$\alpha$ planes from neighboring MPIs. Our MPI layers differ from layered renderings considered in traditional plenoptic sampling because we predict opacities in addition to color for each layer, which allows us to correctly respect occlusions while compositing the depth layer light fields.

In summary, we extend the layered plenoptic sampling framework to correctly handle occlusions by taking advantage of our predicted opacities, and show that this still allows us to increase the required camera sampling interval by a factor of $D$:
\begin{equation}
    \Delta_{u} \leq \frac{D}{2K_{x}f\left(1/z_{\min}-1/z_{\max}\right)}.
\end{equation}

Our framework further differs from classic layered plenoptic sampling in that each MPI is sampled within a reference camera view frustum with a finite field of view, instead of the infinite field of view assumed in prior analyses~\cite{chai00,zhang2003}. In order for the MPI prediction procedure to succeed, every point within the scene's bounding volume should fall within the frustums of at least two neighboring sampled views. The required camera sampling interval $\Delta_{u}$ is then additionally bounded by:
\begin{equation}
\Delta_{u} \leq \frac{W\Delta_{x}z_{\min}}{2f}
\label{eqn:fov}
\end{equation}
where $W$ is the image width in pixels of each sampled view. The overall camera sampling interval must satisfy both constraints:

\begin{equation}
    \Delta_{u} \leq \min\left(\frac{D}{2K_{x}f\left(1/z_{\min}-1/z_{\max}\right)},\frac{W\Delta_{x}z_{\min}}{2f}\right).
\end{equation}

\subsection{Image Space Interpretation of View Sampling}

It is useful to interpret the required camera sampling rate in terms of the maximum pixel disparity $d_{\max}$ of any scene point between adjacent input views. If we set $z_{\max}=\infty$ to allow scenes with content up to an infinite depth and additionally set $K_{x}=1/2\Delta_{x}$ to allow spatial frequencies up to the maximum representable frequency:

\begin{equation}
     \frac{\Delta_{u}f}{\Delta_{x}z_{\min}} = d_{\max} \leq\min\left(D,\frac{W}{2}\right).
    \label{eqn:sampling_req}
\end{equation}

Simply put, the maximum disparity of the closest scene point between adjacent views must be less than $\min(D,W/2)$ pixels. When $D=1$, this inequality reduces to the Nyquist bound: a maximum of 1 pixel of disparity between views.

In summary, promoting each view sample to an MPI scene representation with $D$ depth layers allows us to decrease the required view sampling rate by a factor of $D$, up to the required field of view overlap for stereo geometry estimation. Light fields for real 3D scenes must be sampled in two viewing directions, so this benefit is compounded into a sampling reduction of $D^2$. Section~\ref{sec:sampling_validation} empirically validates that our algorithm's performance matches this theoretical analysis. Section~\ref{sec:sampling_guidelines} describes how we apply the above theory along with the empirical performance of our deep learning pipeline to prescribe practical sampling guidelines for users.

\begin{figure}
\centering
    \includegraphics[width=\columnwidth]{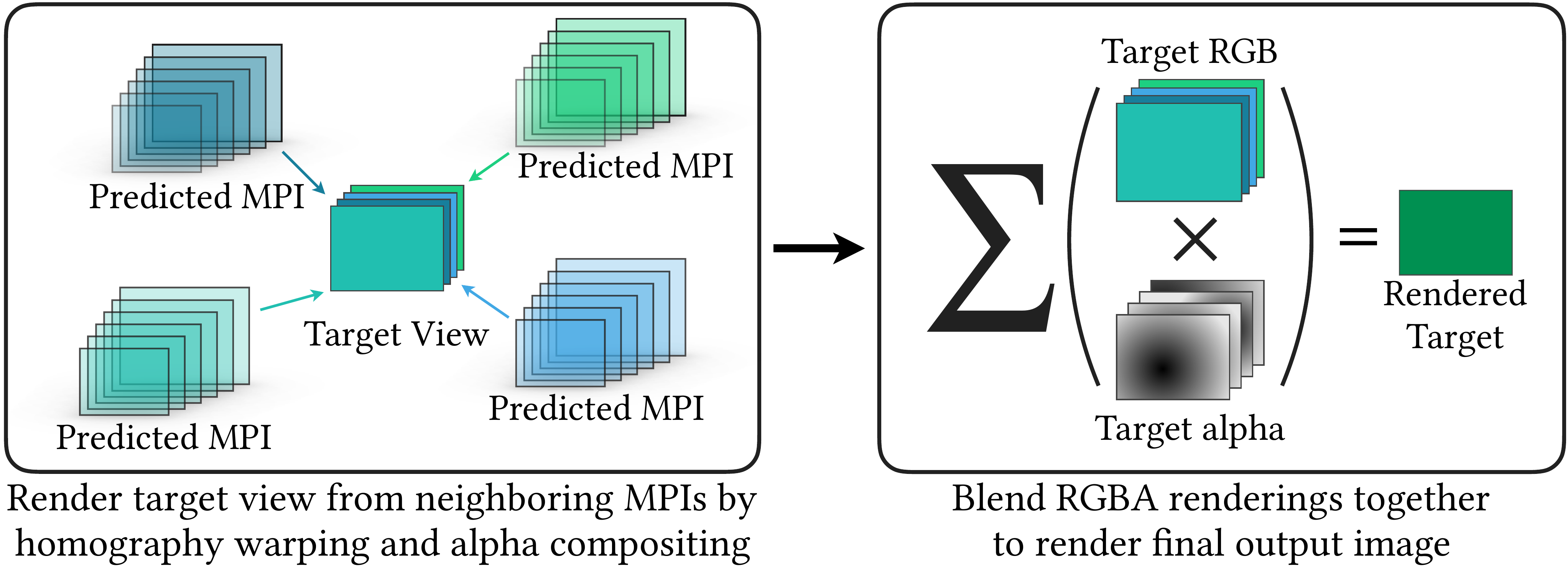}
    \vspace{-3mm}
    \caption{We render novel views as a weighted combination of renderings from neighboring MPIs, modulated by the corresponding accumulated alphas.}
    \label{fig:blend}
\end{figure}

\section{Practical View Synthesis Pipeline}
\label{sec:pipeline}

We present a practical and robust method for synthesizing new views from a set of input images and their camera poses.
Our method first uses a CNN to promote each captured input image to an MPI, then reconstructs novel views by blending renderings from nearby MPIs. Figures~\ref{fig:teaser} and~\ref{fig:blend} visualize this pipeline. We discuss the practical image capture process enabled by our method in Section~\ref{sec:usage}.

\subsection{MPI Prediction for Local Light Field Expansion}
\label{sec:cnn}

The first step in our pipeline is expanding each sampled view to a local light field using an MPI scene representation. 
Our MPI prediction pipeline takes five views as input: the reference view to be expanded and its four nearest neighbors in 3D space.
Each image is reprojected to $D$ depth planes, sampled linearly in disparity within the reference view frustum, to form 5 plane sweep volumes (PSVs) of size $H\times W\times D\times 3$. 

Our 3D CNN takes these 5 PSVs as input, concatenated along the channel dimension. This CNN outputs an opacity $\alpha$ for each MPI coordinate $(x,y,d)$ as well as a set of 5 color selection weights that sum to 1 at each MPI coordinate. These weights parameterize the RGB values in the output MPI as a weighted combination of the input PSVs. Intuitively, each predicted MPI softly ``selects'' its color values at each MPI coordinate from the pixel colors at that coordinate in each of the input PSVs. We specifically use this RGB parameterization instead of the foreground+background parameterization proposed by Zhou \etal~\shortcite{zhou18} because their method does not allow an MPI to directly incorporate content occluded from the reference view but visible in other input views.

Furthermore, we enhance the MPI prediction CNN architecture from the original version to use 3D convolutional layers instead of the original 2D convolutional layers so that our architecture is fully convolutional along the height, width, and depth dimensions. This enables us to predict MPIs with a variable number of planes $D$ so that we can jointly choose the view and disparity sampling densities to satisfy Equation~\ref{eqn:sampling_req}. Table~\ref{table:synth_quant} validates the benefit of being able to change the number of MPI planes to correctly match our derived sampling requirements, enabled by our use of 3D convolutions. Our full network architecture can be found in Appendix~\ref{sec:network_architecture}.

\newcommand{\blendlotrwidth}{.238\columnwidth}
\newcommand{\blendlotrwidthb}{.32\columnwidth}

\begin{figure}
\captionsetup[subfigure]{font=scriptsize,labelformat=empty,aboveskip=1pt,belowskip=2pt}
  \centering
  \begin{subfigure}[t]{\blendlotrwidth}
    \centering\includegraphics[width=\textwidth]{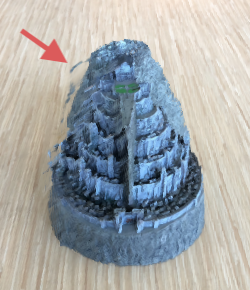}
    \caption{$C_{t,1}$}
  \end{subfigure}
  \begin{subfigure}[t]{\blendlotrwidth}
    \centering\includegraphics[width=\textwidth]{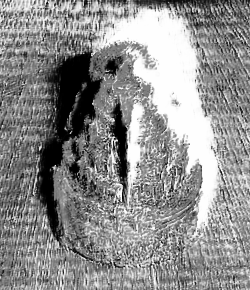}
    \caption{$\alpha_{t,1}/(\alpha_{t,1}+\alpha_{t,2})$}
  \end{subfigure}
  \begin{subfigure}[t]{\blendlotrwidth}
    \centering\includegraphics[width=\textwidth]{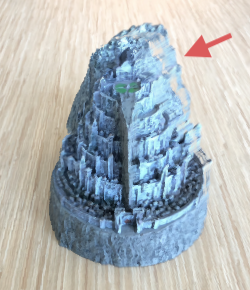}
    \caption{$C_{t,2}$}
  \end{subfigure}
  \begin{subfigure}[t]{\blendlotrwidth}
    \centering\includegraphics[width=\textwidth]{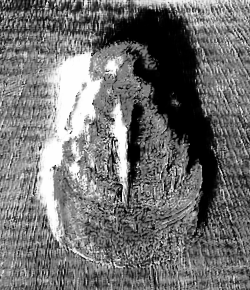}
    \caption{$\alpha_{t,2}/(\alpha_{t,1}+\alpha_{t,2})$}
  \end{subfigure}

  \begin{subfigure}[t]{\blendlotrwidthb}
    \centering\includegraphics[width=\textwidth]{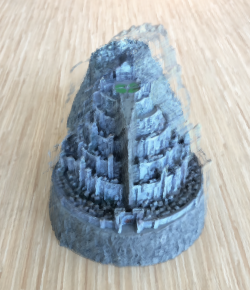}
    \caption{Average of $C_{t,i}$}
  \end{subfigure}
  \begin{subfigure}[t]{\blendlotrwidthb}
    \centering\includegraphics[width=\textwidth]{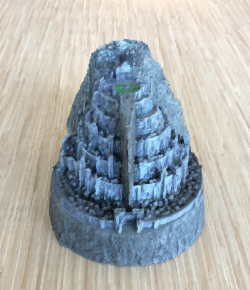}
    \caption{Blended with $\alpha$}
  \end{subfigure}
  \begin{subfigure}[t]{\blendlotrwidthb}
    \centering\includegraphics[width=\textwidth]{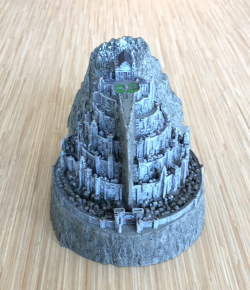}
    \caption{Ground truth}
  \end{subfigure}
\vspace{-3mm}
  \caption{An example illustrating the benefits of using accumulated alpha to blend MPI renderings. We render two MPIs at the same new camera pose. In the top row, we display the RGB outputs $C_{t,i}$ from each MPI as well as the accumulated alphas $\alpha_{t,i}$, normalized so that they sum to one at each pixel. In the bottom row, we see that a simple average of the RGB images $C_{t,i}$ retains the stretching artifacts from both MPI renderings, whereas the alpha weighted blending combines only the non-occluded pixels from each input to produce a clean output $C_t$.}
\label{fig:blend_lotr}

\end{figure}

\subsection{Continuous View Reconstruction by Blending}
\label{sec:blending}

As discussed in Section~\ref{sec:sampling}, we reconstruct interpolated views as a weighted combination of renderings from multiple nearby MPIs. This effectively combines our local light field approximations into a light field with a near plane spanning the extent of the captured input views and a far plane determined by the field-of-view of the input views. 
As in standard light field rendering, this allows for a new view path with unconstrained 3D translation and rotation within the range of views made up of rays in the light field.

One important detail in our rendering process is that we consider the accumulated alpha values from each MPI rendering when blending. This allows each MPI rendering to ``fill in'' content that is occluded from other camera views.

Our MPI prediction network uses a set of RGB images $C_k$ along with their camera poses $p_k$ to produce a set of MPIs $M_k$ (one corresponding to each input image). To render a novel view with pose $p_t$ using the predicted MPI $M_k$, we homography warp each RGB$\alpha$ MPI plane into the frame of reference of the target pose $p_t$ then alpha composite the warped planes together from back to front. This produces an RGB image and an alpha image, which we denote $C_{t,k}$ and $\alpha_{t,k}$ respectively (subscript $t,k$ indicating that the output is rendered at pose $p_t$ using the MPI at pose $p_k$).

Since a single MPI alone will not necessarily contain all the content visible from the new camera pose due to occlusions and field of view issues, we generate the final RGB output $C_{t}$ by blending rendered RGB images $C_{t,k}$ from multiple MPIs, as depicted in Figure~\ref{fig:blend}. We use scalar blending weights $w_{t,k}$, each modulated by the corresponding accumulated alpha images $\alpha_{t,k}$ and normalized so that the resulting rendered image is fully opaque ($\alpha=1$):

\begin{equation}
    C_t=\frac{\sum_k w_{t,k} \alpha_{t,k} C_{t,k}}{\sum_k w_{t,k} \alpha_{t,k}}.
    \label{eqn:alpha_blend}
\end{equation}
For an example where modulating the blending weights by the accumulated alpha values prevents artifacts in $C_t$, see Figure~\ref{fig:blend_lotr}. Table~\ref{table:synth_quant} demonstrates that blending with alpha gives quantitatively superior results over both using a single MPI and blending multiple MPI renderings without using the accumulated alpha.

The blending weights $w_{t,k}$ can be any sufficiently smooth filter. In the case of data sampled on a regular grid, we use bilinear interpolation from the four nearest MPIs rather than the ideal sinc function interpolation for effiency and due to the limited number of sampled views. For irregularly sampled data, we use the five nearest MPIs and take $w_{t,k} \propto \exp\left(-\gamma \ell(p_t,p_k)\right)$. Here $\ell(p_t,p_k)$ is the $L^2$ distance between the translation vectors of poses $p_t$ and $p_k$, and the constant $\gamma$ is defined as $\frac{f}{Dz_{\min}}$ given focal length $f$, minimum distance to the scene $z_{\min}$, and number of planes $D$. (Note that the quantity $\frac{f\ell}{z_{\min}}$ represents $\ell$ converted into units of pixel disparity.)

\newcommand{\platewidth}{.48\columnwidth}
\begin{figure}
\captionsetup[subfigure]{font=scriptsize,labelformat=empty,aboveskip=1pt,belowskip=2pt}
  \centering
  
  \begin{subfigure}[c]{.6\columnwidth}
    \centering\includegraphics[width=\textwidth]{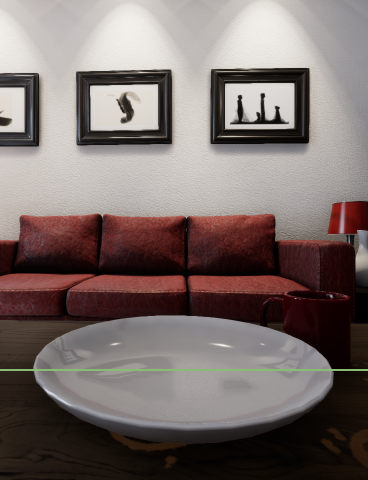}
    \caption{Central image (ground truth)}
  \end{subfigure}
  \begin{subfigure}[c]{.38\columnwidth}
    \centering\includegraphics[width=\textwidth]{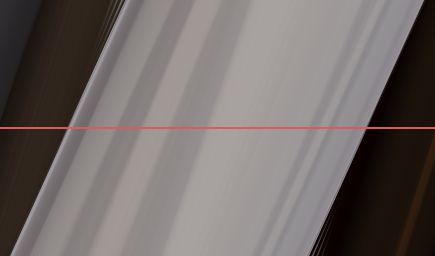}
    \caption{Single MPI}
    \centering\includegraphics[width=\textwidth]{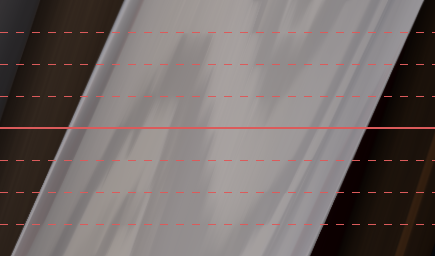}
    \caption{Blended MPIs}
    \centering\includegraphics[width=\textwidth]{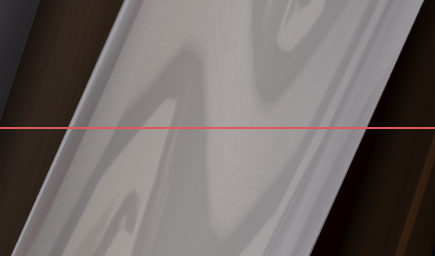}
    \caption{Ground truth}
  \end{subfigure}
  
  \caption{We demonstrate that a collection of MPIs can approximate a highly non-Lambertian light field. In this synthetic scene, the curved plate reflects the paintings on the wall, leading to quickly-varying specularities as the camera moves horizontally. This effect can be observed in the ground truth epipolar plot (bottom right). A single MPI (top right) can only place a specular reflection at a single virtual depth, but blending renderings from multiple MPIs (middle right) provides a much better approximation to the true light field. In this example, we blend between MPIs evenly distributed at every 32 pixels of disparity along a horizontal path, indicated by the dashed lines in the epipolar plot.}
\label{fig:plate_spec}
\end{figure}

Our strategy of blending between neighboring MPIs is particularly effective for rendering non-Lambertian effects. For general curved surfaces, the virtual apparent depth of a specularity changes with the viewpoint~\cite{swaminathan02}. As a result, specularities appear as curves in epipolar slices of the light field, while diffuse points appear as lines. Each of our predicted MPIs can represent a specularity for a local range of views by placing the specularity at a single virtual depth. Figure~\ref{fig:plate_spec} illustrates how our rendering procedure effectively models a specularity's curve in the light field by blending locally linear approximations, as opposed to the limited extrapolation provided by a single MPI.

\section{Training Our View Synthesis Pipeline}

\subsection{Training Dataset}

We train our view synthesis pipeline using both renderings and real images of natural scenes. Using synthetic training data crucially enables us to easily generate a large dataset with input view and scene depth distributions similar to those we expect at test time, while using real data helps us generalize to real-world lighting and reflectance effects as well as small errors in pose estimation. 

Our synthetic training set consists of images rendered from the SUNCG~\cite{suncg} and UnrealCV~\cite{unrealcv} datasets. SUNCG contains 45,000 simplistic house and room environments with texture mapped surfaces and low geometric complexity. UnrealCV contains only a few large scale environments, but they are modeled and rendered with extreme detail, providing geometric complexity, texture variety, and non-Lambertian reflectance effects. We generate views for each synthetic training instance by first randomly sampling a target baseline for the inputs (up to 128 pixels of disparity), then randomly perturbing the camera pose in 3D to approximately match this baseline.

Our real training dataset consists of 24 scenes from our handheld cellphone captures, with 20-30 images each. We use the COLMAP structure from motion~\cite{schoenberger2016sfm} implementation to compute poses for our real images.

\subsection{Training Procedure}

For each training step, we sample two sets of 5 views each to use as inputs, and a single held-out target view for supervision.
We first use the MPI prediction network to predict two MPIs, one from each set of 5 inputs. Next, we render the target novel view from both MPIs and blend these renderings using the accumulated alpha values, as described in Equation~\ref{eqn:alpha_blend}.

The training loss is simply the image reconstruction loss for the rendered novel view. We follow the original work on MPI prediction~\cite{zhou18} and use a VGG network activation perceptual loss as implemented by Chen and Koltun~\shortcite{chen2017}, which has been consistently shown to outperform standard image reconstruction losses~\cite{huang18,lpips}.
We are able to supervise only the final blended rendering because both our fixed rendering and blending functions are differentiable. Learning through this blending step trains our MPI prediction network to leave alpha ``holes'' in uncertain regions for each MPI, in the expectation that this content will be correctly rendered by another neighboring MPI, as illustrated by Figure~\ref{fig:blend_lotr}. 

In practice, training through blending is slower than training a single MPI, so we first train the network to render a new view from only one MPI for 500k iterations, then train the full pipeline (blending views from two different MPIs) for 100k iterations. To fine tune the network to process real data, we train on our small real dataset for an additional 10k iterations. We use $320\times 240$ resolution and up to 128 planes for SUNCG training data, and $640 \times 480$ resolution and up to 32 planes for UnrealCV training data, due to GPU memory limitations. 
We implement our full pipeline in Tensorflow~\cite{tensorflow} and optimize the MPI prediction network parameters using Adam~\cite{adam} with a learning rate of $2\times 10^{-4}$ and a batch size of one. We split the training pipeline across two Nvidia RTX 2080Ti GPUs, using one GPU to generate each MPI.

\begin{figure}
\centering
    \includegraphics[width=\columnwidth]{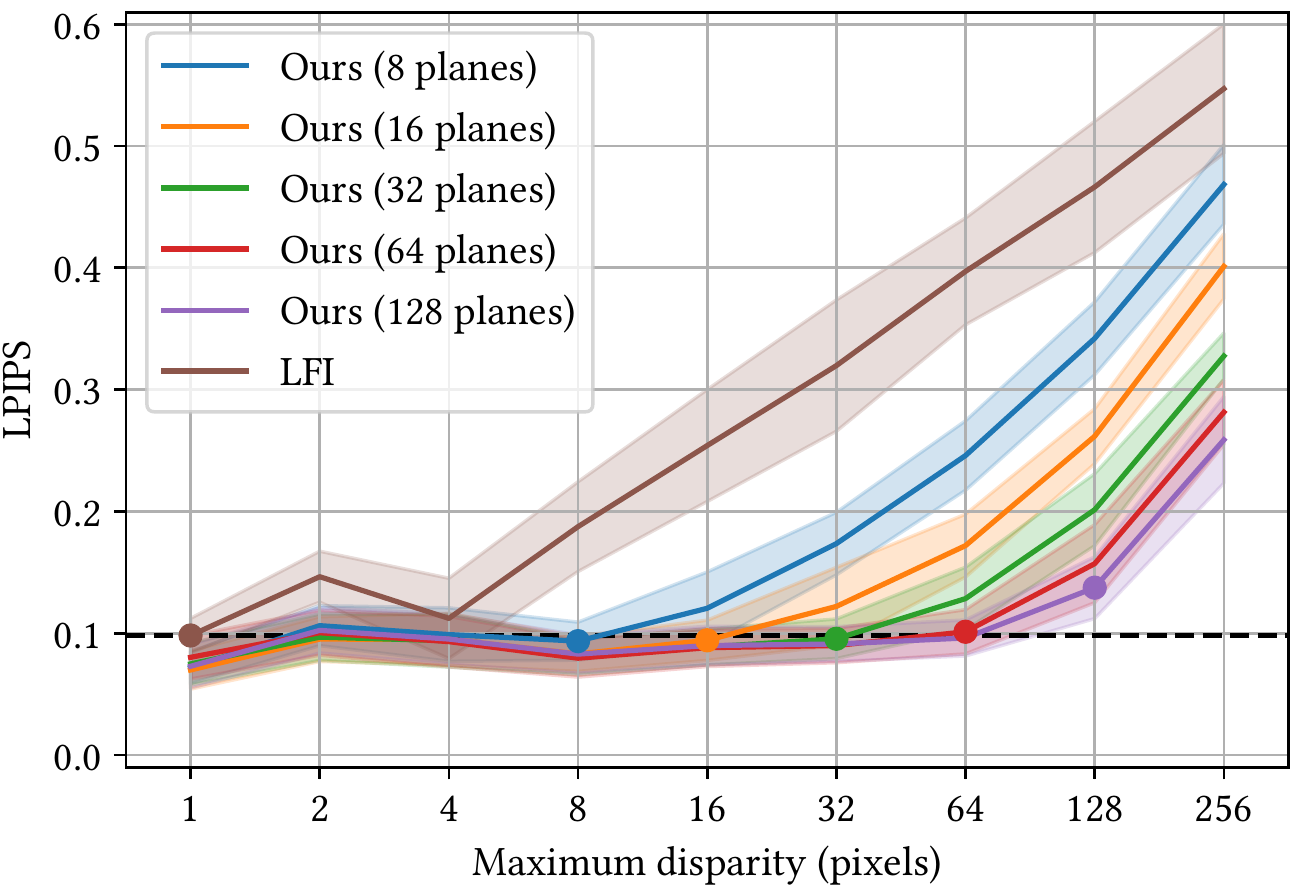}
    \caption{We plot the performance of our method (with varying number of planes $D=8,16,32,64,$ and $128$) compared to light field interpolation for different input view sampling rates (denoted by maximum scene disparity $d_{\max}$ between adjacent input views). Our method can achieve the same perceptual quality as LFI with Nyquist rate sampling (black dotted line) as long as the number of predicted planes matches or exceeds the undersampling rate, up to an undersampling rate of 128. At $D=64$, this means we achieve the same quality as LFI with $64^2\approx 4000\times$ fewer views. We use the LPIPS~\cite{lpips} metric (lower is better) because we primarily value perceptual quality. The colored dots indicate the point on each line where the number of planes equals the maximum scene disparity, where equality is achieved in our sampling bound (Equation~\ref{eqn:sampling_req}). The shaded region indicates $\pm 1$ standard deviation over all 8 test scenes.}
\label{fig:disp_vs_planes}
\end{figure}

\section{Experimental Evaluation}

We quantitatively and qualitatively validate our method's prescriptive sampling benefits and ability to render high fidelity novel views of light fields that have been undersampled by up to $4000\times$, as well as demonstrate that our algorithm outperforms state-of-the-art methods for regular view interpolation. Figure~\ref{fig:real_results} showcases these qualitative comparisons on scenes with complex geometry (Fern and T-Rex) and highly non-Lambertian scenes (Air Plants and Pond) that are not handled well by most view synthesis algorithms. 

For all quantitative comparisons (Table~\ref{table:synth_quant}), we use a synthetic test set rendered from an UnrealCV~\cite{unrealcv} environment that was not used to generate any training data. Our test set contains 8 scenes, each rendered at $640\times 480$ resolution and at 8 different view sampling densities such that the maximum disparity between adjacent input views ranges from 1 to 256 pixels (a maximum disparity of 1 pixel between input views corresponds to Nyquist rate view sampling). We restrict our quantitative comparisons to rendered images because a Nyquist rate grid-sampled light field would require at least $384^2$ camera views to generate a similar test set, and no such densely-sampled real light field dataset exists to the best of our knowledge. We report quantitative performance using the standard PSNR and SSIM metrics, as well as the state-of-the-art LPIPS~\cite{lpips} perceptual metric, which is based on a weighted combination of neural network activations tuned to match human judgements of image similarity.

Finally, our accompanying video shows results on over 60 additional real-world scenes. These renderings were created completely automatically by a script that takes only the set of captured images and desired output view path as inputs, highlighting the practicality and robustness of our method.

\subsection{Sampling Theory Validation}
\label{sec:sampling_validation}

Our method is able to render high-quality novel views while significantly decreasing the required input view sampling density compared to standard light field interpolation. Figure~\ref{fig:disp_vs_planes} shows that our method is able to render novel views with Nyquist level perceptual quality with up to $d_{\max}=64$ pixels of disparity between input view samples, as long as we match the number of planes in each MPI to the maximum pixel disparity between input views. We postulate that our inability to match Nyquist quality from input images with a maximum of 128 pixels of disparity is due to the effect of occlusions. It becomes increasingly likely that any non-foreground scene point will be sampled by fewer input views as the maximum disparity between adjacent views increases. This increases the difficulty of depth estimation and requires the CNN to hallucinate the appearance and depth of occluded points in extreme cases where they are sampled by none of the input views.

Figure~\ref{fig:disp_vs_planes} also shows that once our sampling bound is satisfied, adding additional planes does not increase performance. For example, at 32 pixels of disparity, increasing from 8 to 16 to 32 planes decreases the LPIPS error, but performance stays constant from 32 to 128 planes. This verifies that for scenes up to 64 pixels of disparity, adding additional planes past the maximum pixel disparity between input views is of limited value, in accordance with our theoretical claim that partitioning a scene with disparity variation of $D$ pixels into $D$ depth ranges is sufficient for continuous reconstruction.

\newcommand{\bestnum}[1]{$\mathbf{#1}$}
\newcommand{\secnum}[1]{$\mathit{#1}$}

\begin{table*}
\begin{center}
\caption{
We quantitatively show that our method outperforms state-of-the-art baselines and specific ablations of our method, across a wide range of input sampling rates (measured by the maximum pixel disparity $d_{\max}$ between adjacent input views), on a synthetic test set. We display results using the standard PSNR and SSIM metrics (higher is better) as well as the LPIPS perceptual metric~\cite{lpips} (lower is better). The best measurement in each column is bolded. See Sections~\ref{sec:baselines} and~\ref{sec:ablation} for details on each comparison.
}
\resizebox{\textwidth}{!}{
\begin{tabular}{ l | l | c c c | c c c | c c c | c c c }
\multicolumn{2}{c}{} &
\multicolumn{12}{c}{Maximum disparity $d_{\max}$ (pixels)} \\
\multicolumn{2}{c|}{}  &
\multicolumn{3}{c|}{16} &
\multicolumn{3}{c|}{32} &
\multicolumn{3}{c|}{64} &
\multicolumn{3}{c}{128} \\
& Algorithm & PSNR $\uparrow$ & SSIM $\uparrow$ & LPIPS $\downarrow$ & PSNR $\uparrow$ & SSIM $\uparrow$ & LPIPS $\downarrow$  & PSNR $\uparrow$ & SSIM $\uparrow$ & LPIPS $\downarrow$  & PSNR $\uparrow$ & SSIM $\uparrow$ & LPIPS $\downarrow$  \\
\hline
 \multirow{4}{*}{Baselines}
& LFI               	 & 26.21 & 0.7776 & 0.2541 & 23.35 & 0.6982 & 0.3198 & 20.60 & 0.6243 & 0.3971 & 18.32 & 0.5560 & 0.4665 \\
& ULR               	 & 28.17 & 0.8320 & 0.1510 & 26.43 & 0.7987 & 0.1820 & 24.34 & 0.7679 & 0.2311 & 21.24 & 0.7062 & 0.3215 \\
& Soft3D             	 & 34.48 & 0.9430 & 0.1345 & 32.33 & 0.9216 & 0.1795 & 27.97 & 0.8588 & 0.2652 & 23.11 & 0.7382 & 0.3979 \\
& BW Deep           	 & 34.18 & 0.9433 & 0.1074 & 34.00 & 0.9476 & 0.1128 & 31.88 & 0.9192 & 0.1573 & 27.59 & 0.8363 & 0.2591 \\
\hline
\multirow{2}{*}{Ablations} 
& Single MPI           	 & 31.11 & 0.9482 & 0.1007 & 29.38 & 0.9424 & 0.1111 & 26.88 & 0.9250 & 0.1363 & 24.20 & 0.8734 & 0.1980 \\
& Avg. MPIs           	 & 32.67 & 0.9560 & 0.1140 & 31.34 & 0.9532 & 0.1248 & 29.31 & 0.9400 & 0.1423 & 27.02 & 0.8999 & 0.1961 \\
\hline
& Ours              	 & \bestnum{34.57} & \bestnum{0.9568} & \bestnum{0.0942} & \bestnum{34.48} & \bestnum{0.9569} & \bestnum{0.0954} & \bestnum{33.58} & \bestnum{0.9530} & \bestnum{0.1012} & \bestnum{31.96} & \bestnum{0.9323} & \bestnum{0.1374 }
\end{tabular}
}
\label{table:synth_quant}
\end{center}
\end{table*}

\subsection{Comparisons to Baseline Methods}
\label{sec:baselines}

We quantitatively (Table~\ref{table:synth_quant}) and qualitatively (Figure~\ref{fig:real_results}) demonstrate that our algorithm produces superior renderings, particularly for non-Lambertian effects, without the artifacts seen in renderings from competing methods. We urge readers to view our accompanying video for convincing rendered camera paths that highlight the benefits of our approach.

We compare our method to state-of-the-art view synthesis techniques as well as view-dependent texture mapping using a global mesh as proxy geometry. Please refer to Appendix~\ref{sec:baseline_implementation} for additional implementation details regarding baseline methods.

\paragraph{Light Field Interpolation (LFI)~\cite{chai00}} This baseline is representative of continuous view reconstruction based on classic signal processing. Following the method of plenoptic sampling~\cite{chai00}, we render novel views using a bilinear interpolation reconstruction filter sheared to the mean scene disparity. Figure~\ref{fig:real_results} demonstrates that increasing the camera spacing beyond the Nyquist rate results in aliasing and ghosting artifacts when using this method.

\paragraph{Unstructured Lumigraph Rendering (ULR)~\cite{buehler01}} This baseline is representative of view dependent texture mapping with an estimated global mesh as a geometry proxy. We reconstruct a global mesh from all inputs using the screened Poisson surface reconstruction algorithm~\cite{kazhdan13}, and use the heuristic Unstructured Lumigraph blending weights~\cite{buehler01} to blend input images after reprojecting them into the novel viewpoint using the global mesh. We use a plane at the mean scene disparity as a proxy geometry to fill in holes in the mesh.

It is particularly difficult to reconstruct a global mesh with geometry edges that are well-aligned with image edges, which causes perceptually jarring artifacts. Furthermore, mesh reconstruction often fails to fill in large portions of the scene, resulting in ghosting artifacts similar to those seen in light field interpolation.

\paragraph{Soft3D~\cite{penner17}} Soft3D is a state-of-the-art view synthesis algorithm that is similar to our approach in that it also computes a local layered scene representation for each input view and projects and blends these volumes to render each novel view. However, it uses a hand-crafted pipeline based on classic local stereo and guided filtering to compute each layered representation. Furthermore, since classic stereo methods are unreliable for smooth or repetitive image textures and non-Lambertian materials, Soft3D relies on smoothing their geometry estimation across many (up to 25) input views. 

Table~\ref{table:synth_quant} quantitatively demonstrates that our approach outperforms Soft3D overall. In particular, Soft3D's performance degrades much more rapidly as the input view sampling rate decreases since their aggregation is less effective when fewer input images view the same scene content. Our method is able to predict high-quality geometry in scenarios where Soft3D suffers from noisy and erroneous results of local stereo because we leverage deep learning to learn implicit priors on natural scene geometry. This is in line with recent work that has shown the benefits of deep learning over traditional stereo for depth estimation~\cite{kendall2017,huang18}.

Figure~\ref{fig:real_results} qualitatively demonstrates that Soft3D generally contains blurred geometry artifacts due to errors in local depth estimation, and that Soft3D's approach fails for rendering non-Lambertian effects because their aggregation procedure blurs the specularity geometry, which changes with the input image viewpoint. 

\paragraph{Backwards warping deep network (BW Deep)} This baseline subsumes recent deep learning view synthesis techniques~\cite{kalantari16, flynn16}, which use a CNN to estimate geometry for each novel view and then backwards warp and blend nearby input images to render the target view. We train a network that uses the same 3D CNN architecture as our MPI prediction network but instead outputs a single depth map at the pose of the new target view. We then backwards warp the five input images into the new view using this depth map and use a second 2D CNN to composite these warped input images into a single rendered output view. 
As shown in Table~\ref{table:synth_quant}, performance for this method degrades quickly as the maximum disparity increases. Although this approach produces comparable images to our method for scenes with small disparities ($d_{\max}=16,32$), the renderings suffer from extreme inconsistency when rendering video sequences.

BW Deep methods use a CNN to estimate depth separately for each output viewpoint, so artifacts appear and disappear over only a few frames, resulting in rapid flickers and pops in the rendered camera path. This inconsistency is visible as corruption in the epipolar plots in Figure~\ref{fig:real_results} and can be clearly seen in our supplemental video. Furthermore, backwards warping incentivizes incorrect depth predictions to fill in disocclusions, so BW Deep methods also produce errors around thin structures and occlusion edges.

\subsection{Ablation Studies}
\label{sec:ablation}

We validate our overall strategy of blending between multiple MPIs as well as our specific blending procedure using accumulated alphas with the following ablation studies:

\paragraph{Single MPI} The fifth row of Table~\ref{table:synth_quant} shows that using only one MPI to produce new views results in significantly decreased performance due to the limited field of view represented in a single MPI as well as depth discretization artifacts as the target view moves far from the MPI reference viewpoint. Additionally, Figure~\ref{fig:plate_spec} shows an example of complex non-Lambertian reflectance that cannot be represented by a single MPI. This ablation can be considered an upper bound on the performance of Zhou \etal~\shortcite{zhou18}, since we use one MPI generated by a higher capacity 3D CNN.

\paragraph{Average MPIs} The sixth row of Table~\ref{table:synth_quant} shows that blending multiple MPI outputs for each novel view without using the accumulated alpha channels results in decreased performance.  Figure~\ref{fig:blend_lotr} visualizes that this simple blending leads to ghosting in regions that are occluded from the poses of any of the MPIs used for rendering, because they will contain incorrect content in disoccluded regions.

\newcommand{\realcompwidthA}{.15\textwidth}
\newcommand{\realcompwidthBA}{.2\textwidth}
\newcommand{\realcompwidthCa}{.144\textwidth}

\begin{figure*}
\captionsetup[subfigure]{font=scriptsize,labelformat=empty,aboveskip=1pt,belowskip=2pt}
  \centering
  
  \begin{subfigure}[t]{\realcompwidthBA}
    \centering\includegraphics[width=\textwidth]{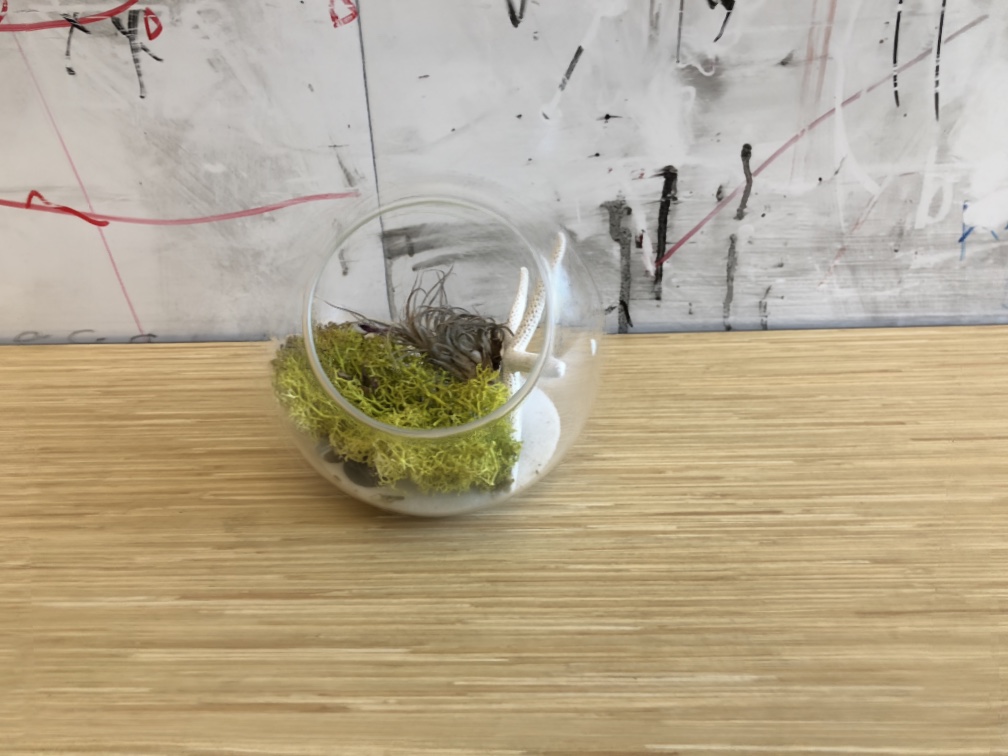}
    \caption{Whole scene (Air Plants)}
    \centering\includegraphics[width=.48\textwidth]{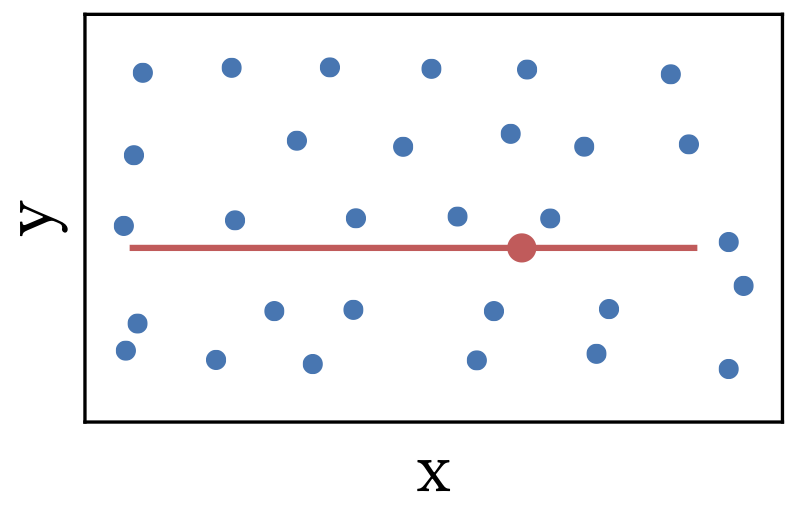}
    \centering\includegraphics[width=.48\textwidth]{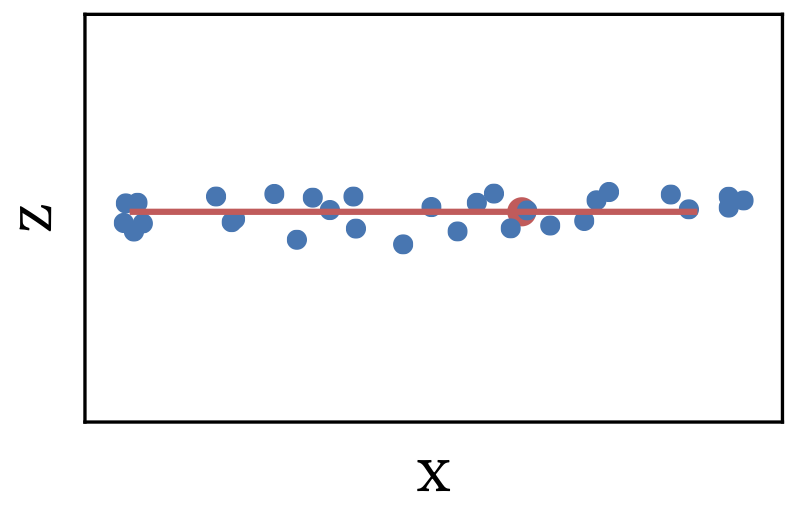}
    \caption{New view path}
  \end{subfigure}
  \begin{subfigure}[t]{\realcompwidthA}
    \centering\includegraphics[width=\textwidth]{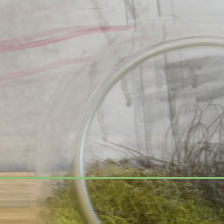}
    \centering\includegraphics[width=\textwidth]{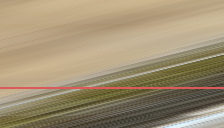}
    \caption{LFI}
  \end{subfigure}
  \begin{subfigure}[t]{\realcompwidthA}
    \centering\includegraphics[width=\textwidth]{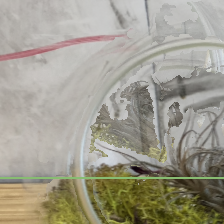}
    \centering\includegraphics[width=\textwidth]{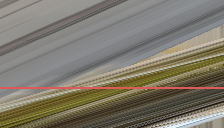}
    \caption{ULR}
  \end{subfigure}
  \begin{subfigure}[t]{\realcompwidthA}
    \centering\includegraphics[width=\textwidth]{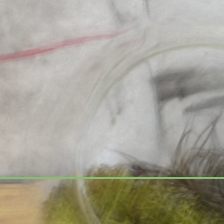}
    \centering\includegraphics[width=\textwidth]{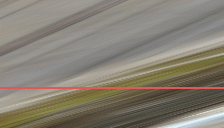}
    \caption{Soft3D}
  \end{subfigure}
  \begin{subfigure}[t]{\realcompwidthA}
    \centering\includegraphics[width=\textwidth]{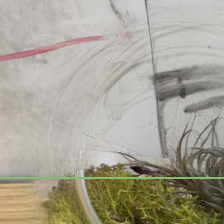}
    \centering\includegraphics[width=\textwidth]{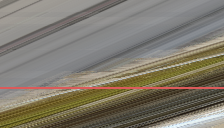}
    \caption{BW Deep}
  \end{subfigure}
  \begin{subfigure}[t]{\realcompwidthA}
    \centering\includegraphics[width=\textwidth]{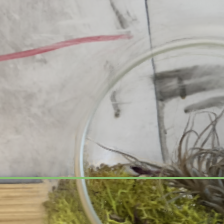}
    \centering\includegraphics[width=\textwidth]{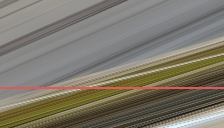}
    \caption{Blended MPIs (ours)}
  \end{subfigure}
  
  \vspace{-0.05in}
  
  \begin{subfigure}[t]{\realcompwidthBA}
    \centering\includegraphics[width=\textwidth]{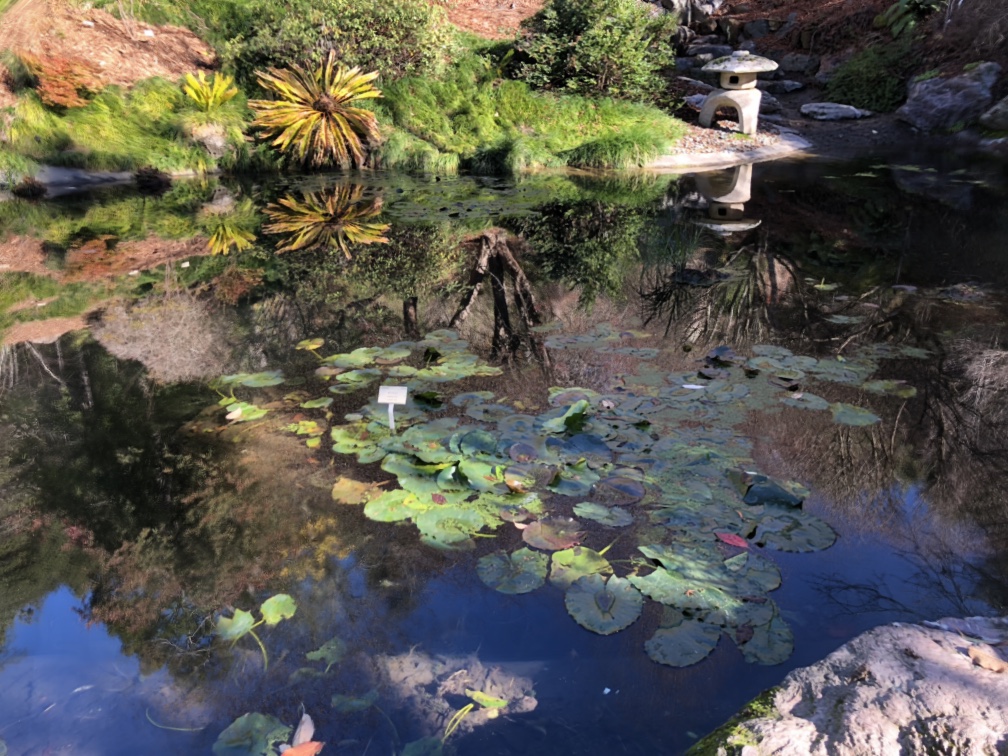}
    \caption{Whole scene (Pond)}
    \centering\includegraphics[width=.48\textwidth]{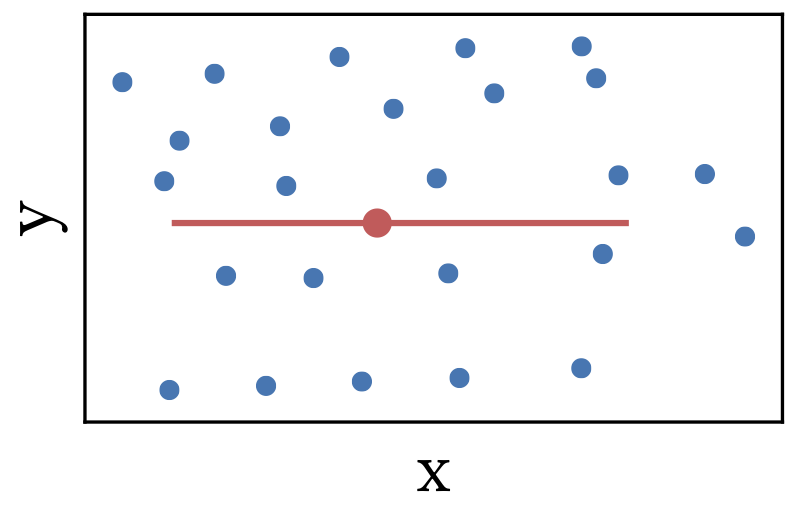}
    \centering\includegraphics[width=.48\textwidth]{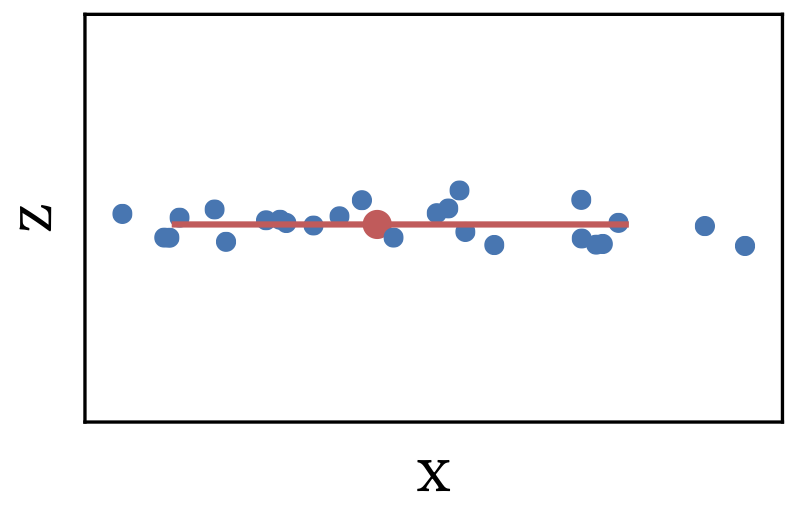}
    \caption{New view path}
  \end{subfigure}
  \begin{subfigure}[t]{\realcompwidthA}
    \centering\includegraphics[width=\textwidth]{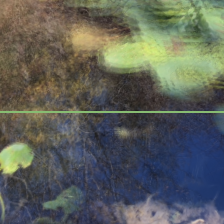}
    \centering\includegraphics[width=\textwidth]{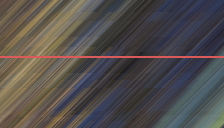}
    \caption{LFI}
  \end{subfigure}
  \begin{subfigure}[t]{\realcompwidthA}
    \centering\includegraphics[width=\textwidth]{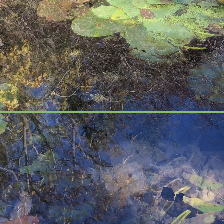}
    \centering\includegraphics[width=\textwidth]{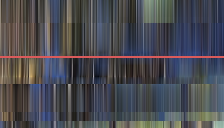}
    \caption{ULR}
  \end{subfigure}
  \begin{subfigure}[t]{\realcompwidthA}
    \centering\includegraphics[width=\textwidth]{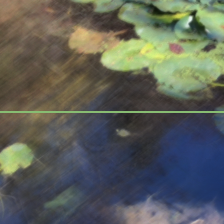}
    \centering\includegraphics[width=\textwidth]{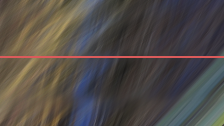}
    \caption{Soft3D}
  \end{subfigure}
  \begin{subfigure}[t]{\realcompwidthA}
    \centering\includegraphics[width=\textwidth]{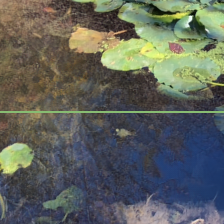}
    \centering\includegraphics[width=\textwidth]{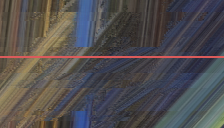}
    \caption{BW Deep}
  \end{subfigure}
  \begin{subfigure}[t]{\realcompwidthA}
    \centering\includegraphics[width=\textwidth]{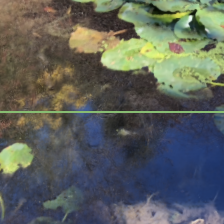}
    \centering\includegraphics[width=\textwidth]{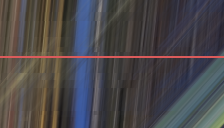}
    \caption{Blended MPIs (ours)}
  \end{subfigure}
  
  \vspace{-0.05in}
  
  \begin{subfigure}[t]{\realcompwidthBA}
    \centering\includegraphics[width=\textwidth]{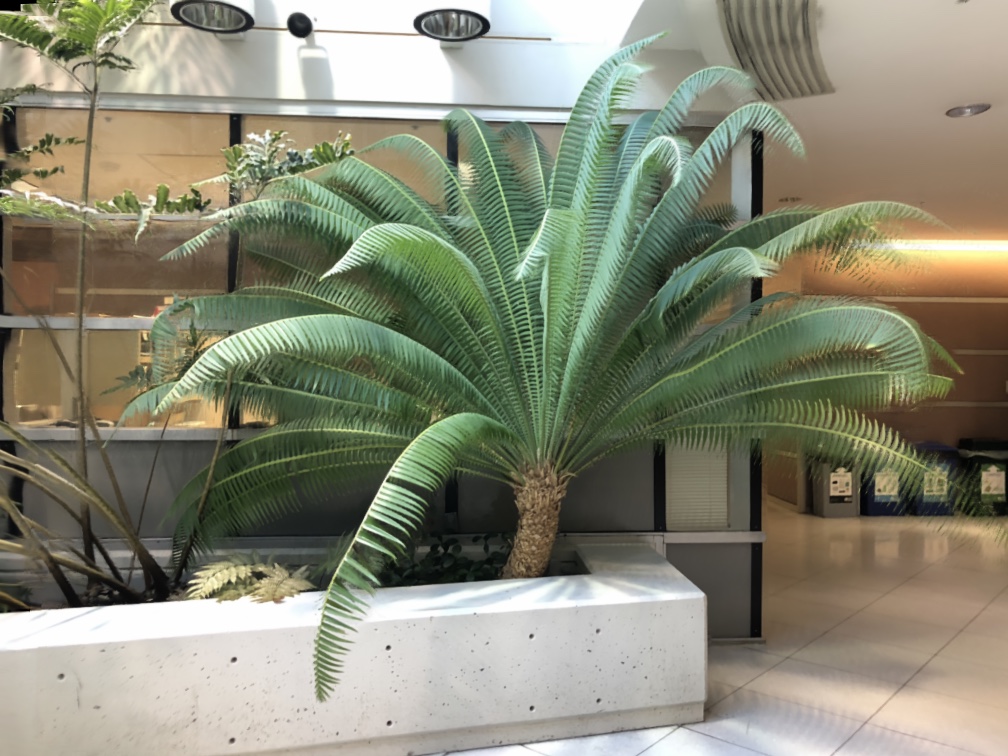}
    \caption{Whole scene (Fern)}
    \centering\includegraphics[width=.48\textwidth]{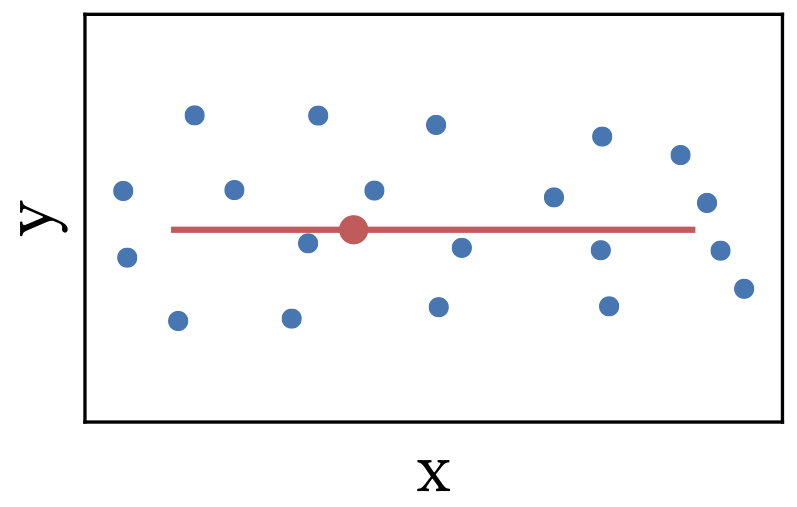}
    \centering\includegraphics[width=.48\textwidth]{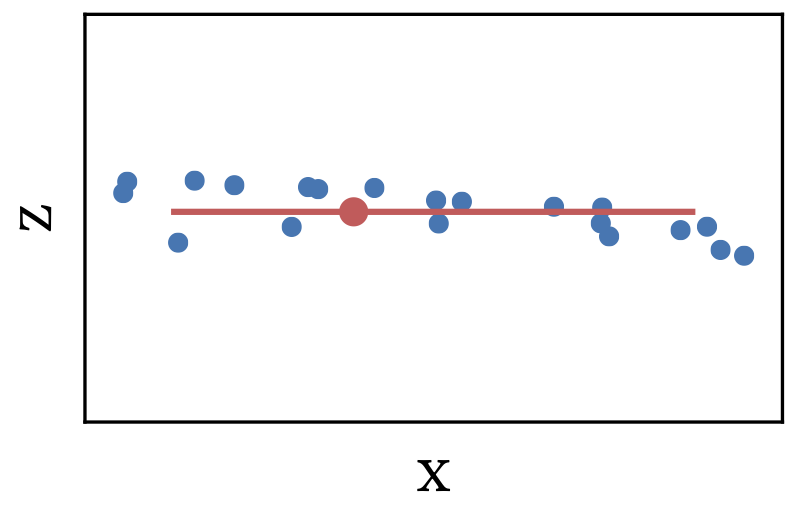}
    \caption{New view path}
  \end{subfigure}
  \begin{subfigure}[t]{\realcompwidthA}
    \centering\includegraphics[width=\textwidth]{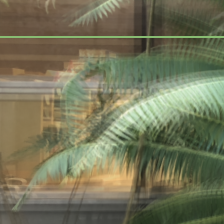}
    \centering\includegraphics[width=\textwidth]{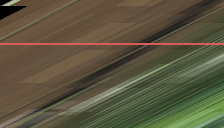}
    \caption{LFI}
  \end{subfigure}
  \begin{subfigure}[t]{\realcompwidthA}
    \centering\includegraphics[width=\textwidth]{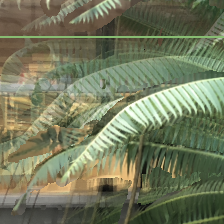}
    \centering\includegraphics[width=\textwidth]{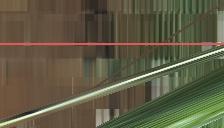}
    \caption{ULR}
  \end{subfigure}
  \begin{subfigure}[t]{\realcompwidthA}
    \centering\includegraphics[width=\textwidth]{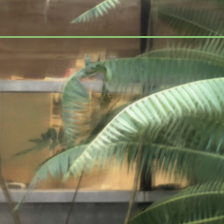}
    \centering\includegraphics[width=\textwidth]{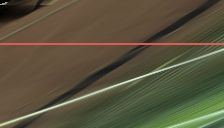}
    \caption{Soft3D}
  \end{subfigure}
  \begin{subfigure}[t]{\realcompwidthA}
    \centering\includegraphics[width=\textwidth]{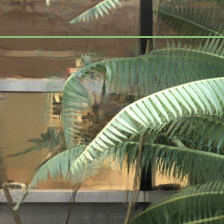}
    \centering\includegraphics[width=\textwidth]{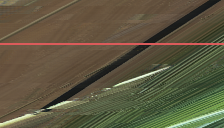}
    \caption{BW Deep}
  \end{subfigure}
  \begin{subfigure}[t]{\realcompwidthA}
    \centering\includegraphics[width=\textwidth]{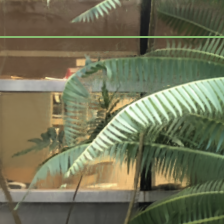}
    \centering\includegraphics[width=\textwidth]{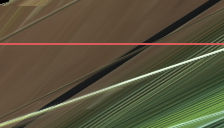}
    \caption{Blended MPIs (ours)}
  \end{subfigure}
  
  \vspace{-0.05in}
  
  \begin{subfigure}[t]{\realcompwidthBA}
    \centering\includegraphics[width=\textwidth]{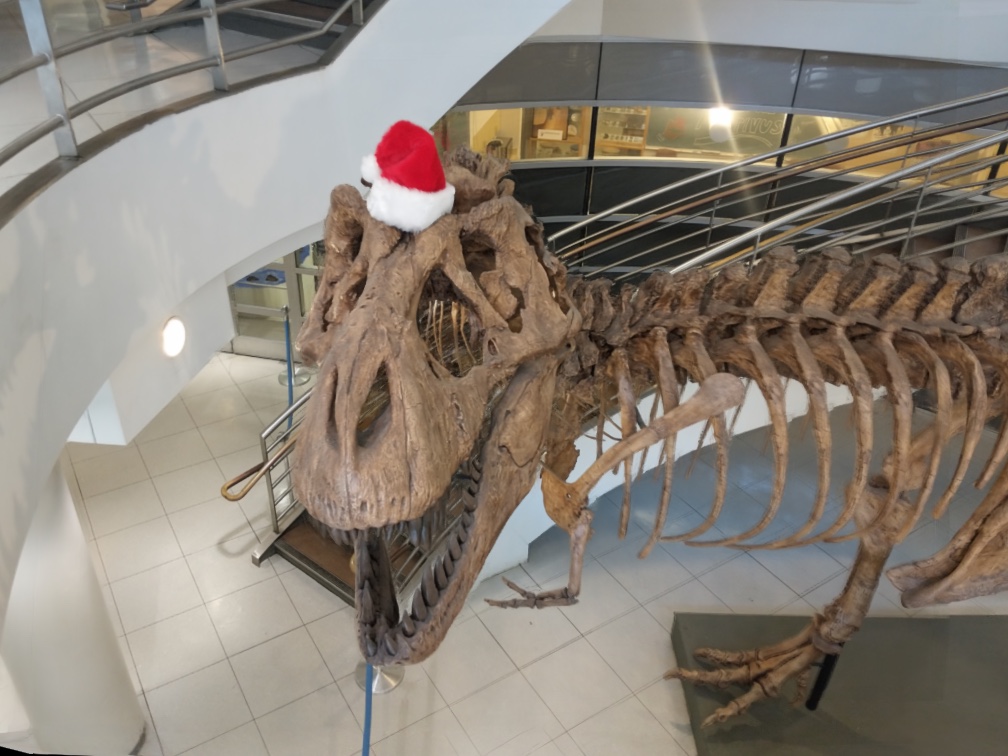}
    \caption{Whole scene (T-Rex)}
    \centering\includegraphics[width=.48\textwidth]{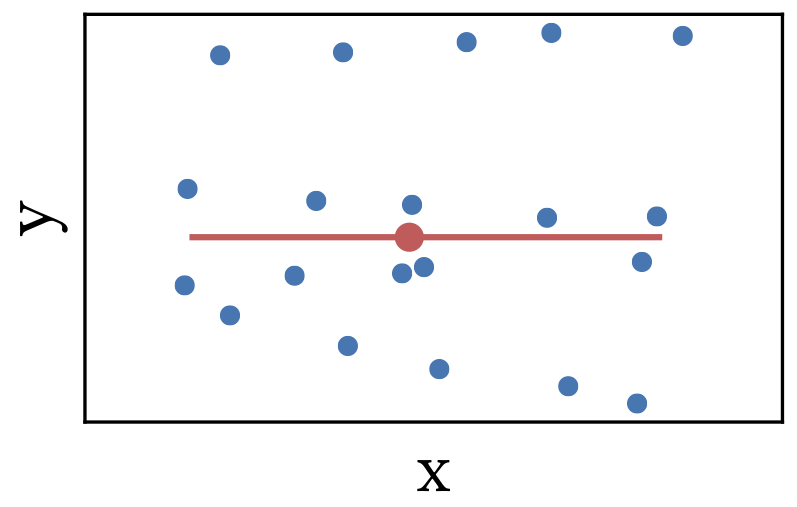}
    \centering\includegraphics[width=.48\textwidth]{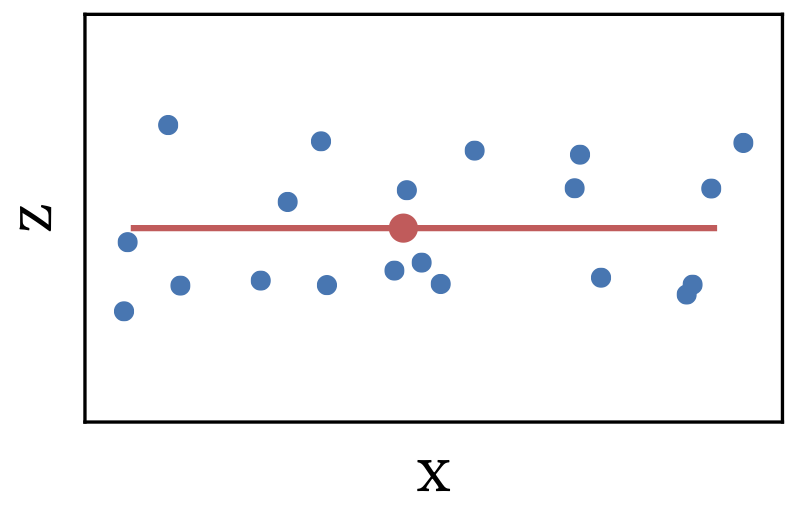}
    \caption{New view path}
  \end{subfigure}
  \begin{subfigure}[t]{\realcompwidthA}
    \centering\includegraphics[width=\textwidth]{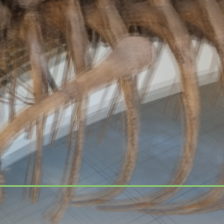}
    \centering\includegraphics[width=\textwidth]{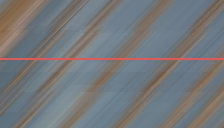}
    \caption{LFI}
  \end{subfigure}
  \begin{subfigure}[t]{\realcompwidthA}
    \centering\includegraphics[width=\textwidth]{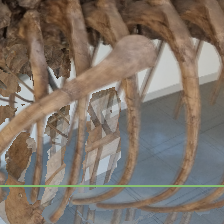}
    \centering\includegraphics[width=\textwidth]{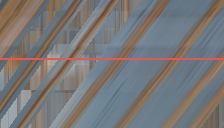}
    \caption{ULR}
  \end{subfigure}
  \begin{subfigure}[t]{\realcompwidthA}
    \centering\includegraphics[width=\textwidth]{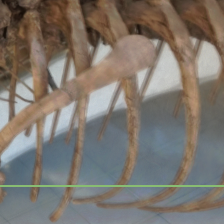}
    \centering\includegraphics[width=\textwidth]{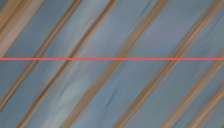}
    \caption{Soft3D}
  \end{subfigure}
  \begin{subfigure}[t]{\realcompwidthA}
    \centering\includegraphics[width=\textwidth]{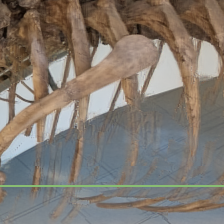}
    \centering\includegraphics[width=\textwidth]{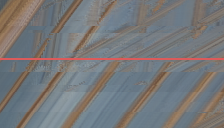}
    \caption{BW Deep}
  \end{subfigure}
  \begin{subfigure}[t]{\realcompwidthA}
    \centering\includegraphics[width=\textwidth]{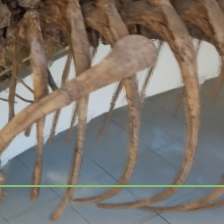}
    \centering\includegraphics[width=\textwidth]{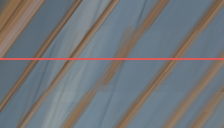}
    \caption{Blended MPIs (ours)}
  \end{subfigure}

\vspace{-3mm}
  \caption{Results on real cellphone datasets. We render a sequence of new views and show both a crop from a single rendered output and an epipolar slice of the sequence. We show 2D projections of the input camera poses (blue dots) and new view path (red line) along the $z$ and $y$ axes of the new view camera in the lower left of each row. LFI fails to cleanly represent objects at different depths because it only uses a single depth plane for reprojection, leading to ghosting (leaves in Fern, lily pads in Pond) and depth inconsistency visible in all epipolar images. Mesh reconstruction failures cause artifacts visible in both the crops and epipolar images for ULR. Soft3D's depth uncertainty leads to blur, and geometry aggregation across large view neighborhoods results in incorrect specularity geometry (brown and blue reflections in Pond). BW Deep's use of a CNN to render every novel view causes depth inconsistency, visible as choppiness across the rows of the epipolar images in all examples. Additionally, BW Deep selects a single depth per pixel, leading to errors for transparencies (glass rim in Air Plants) and reflections (Pond). BW Deep also uses backwards warping, which causes errors around occlusion boundaries (thin ribs in T-Rex). We urge the reader to refer to our supplemental video for high quality videos of these rendered camera paths and additional discussion.}
\label{fig:real_results}

\end{figure*}

\section{Practical Usage}
\label{sec:usage}

We present guidelines to assist users in sampling views that enable high-quality view interpolation with our algorithm, and showcase our method's practicality with a smartphone camera app that guides users to easily capture such input images. Furthermore, we implement a mobile viewer that renders novel views from our predicted MPIs in real-time. Figure~\ref{fig:real_results} showcases examples of rendered results from handheld smartphone captures. Our accompanying video contains a screen capture of our app in use, as well as results on over 60 real-world scenes generated by an automated script.

\subsection{Prescriptive Scene Sampling Guidelines}
\label{sec:sampling_guidelines}

In a typical capture scenario, a user will have a camera with a field of view $\theta$ and a world space plane with side length $S$ that bounds the viewpoints they wish to render. Based on this, we prescribe the design space of image resolution $W$ and number of images to sample $N$ that users can select from to reliably render novel views at Nyquist-level perceptual quality.

Section~\ref{sec:sampling_validation} shows that the empirical limit on the maximum disparity $d_{\max}$ between adjacent input views for our deep learning pipeline is 64 pixels. Substituting Equation~\ref{eqn:sampling_req}:

\begin{equation}
\frac{\Delta_u f}{\Delta_x z_{\min}} \leq 64.
\end{equation}

We translate this into user-friendly quantities by noting that $\Delta_u=S/\sqrt{N}$ and that the ratio of sensor width to focal length $W\Delta_{x}/f=2\tan\theta/2$:

\begin{equation}
\begin{split}
\label{eqn:prescription}
\frac{W}{\sqrt{N}} &\leq \frac{128 z_{\min} \tan(\theta/2)}{S}.
\end{split}
\end{equation}

Using a smartphone camera with a $64^{\circ}$ field of view, this is simply:

\begin{equation}
\label{eqn:smartphone_guideline}
\frac{W}{\sqrt{N}} \leq \frac{80 z_{\min}}{S}.
\end{equation}

Intuitively, once a user has determined the extent of viewpoints they wish to render and the depth of the closest scene point, they can choose any target rendering resolution $W$ and number of images to capture $N$ such that the ratio $W/\sqrt{N}$ satisfies the above expression.

\subsection{Asymptotic Rendering Time and Space Complexity}

Within the possible choices of rendering resolution $W$ and number of sampled views $N$ that satisfy the above guideline, different users may value capture time, rendering time, and storage costs differently. We derive the asymptotic complexities of these quantities to further assist users in choosing correct parameters for their application.

First, the capture time is simply $O(N)$. The render time of each MPI generated is proportional to the number of planes times the pixels per plane:
\begin{equation}
\label{eq:time_complexity}
    W^2D = \frac{W^{3}S}{2\sqrt{N}z_{\min}\tan(\theta/2)} = O(W^3 N^{-1/2}).
\end{equation}
Note that the rendering time for each MPI decreases as the number of sampled images $N$ increases, because this allows us to use fewer planes per MPI. The total MPI storage cost is proportional to: 
\begin{equation}
\label{eq:space_complexity}
    W^2D \cdot N = \frac{W^{3}S\sqrt{N}}{2z_{\min}\tan(\theta/2)} = O(W^3 N^{1/2}).
\end{equation}

\begin{figure}
    \includegraphics[width=\columnwidth]{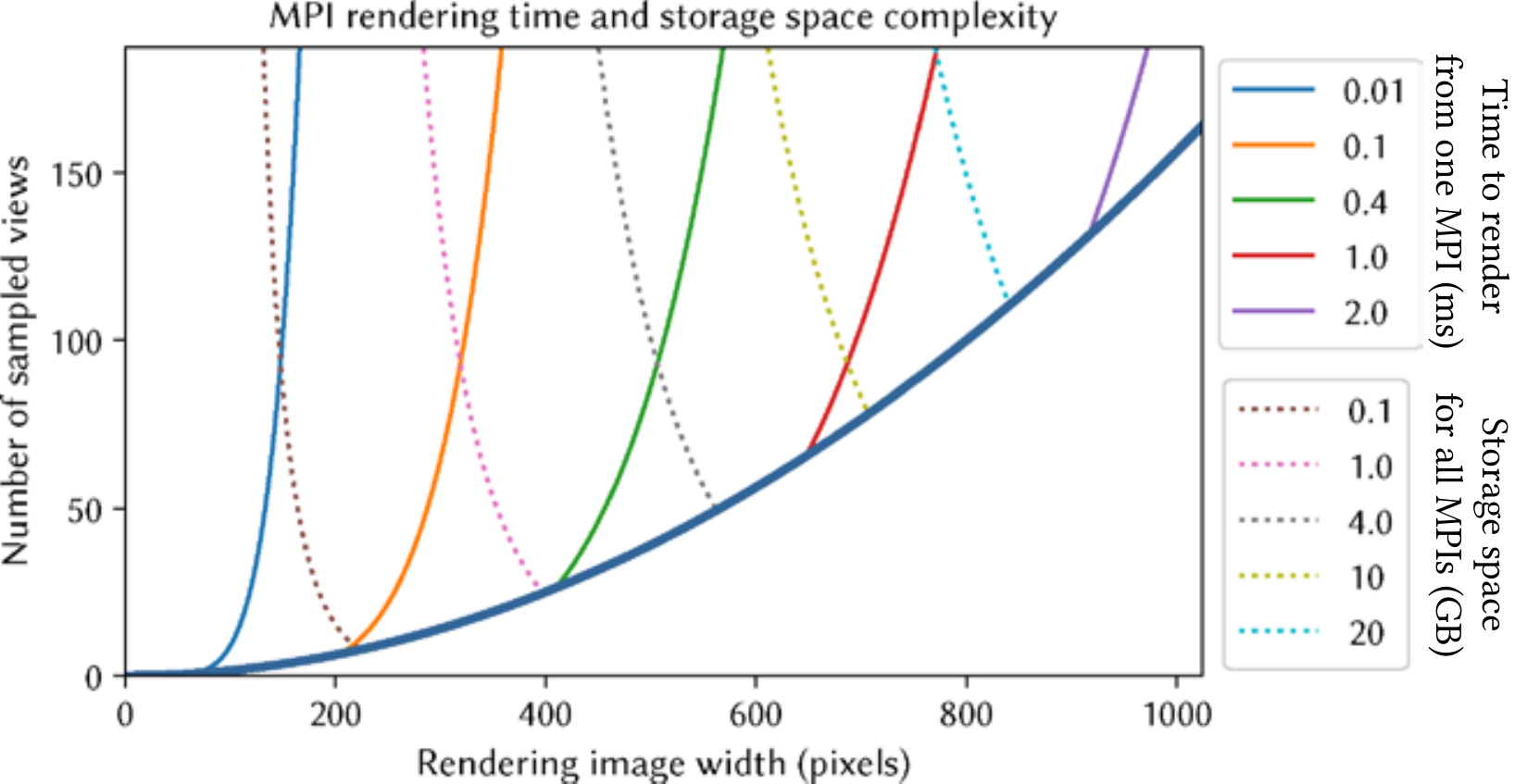}
  \caption{Time and storage cost tradeoff within the space of rendering resolution and number of sampled views that result in Nyquist level perceptual quality (space above the thick blue curve signifying $D=d_{\max} \leq 64$, as in Equation~\ref{eqn:smartphone_guideline}). We plot isocontours of rendering time and storage space for an example scene with close depth $z_{\min}=1.0m$ and target view plane with side length $0.5m$, captured with a camera with a $64^{\circ}$ field of view. We use the average rendering speed from our desktop viewer and the storage requirement from uncompressed 8-bit MPIs. Users can select the point where their desired rendering speed and storage space isocontours intersect to determine the minimum required number of views and maximum affordable rendering resolution.}
\label{fig:tradeoff}

\end{figure}

Practically, this means that users should determine their specific rendering time and storage constraints, and then maximize the image resolution and number of sampled views that satisfy their constraints as well as the guideline in Equation~\ref{eqn:prescription}. Figure~\ref{fig:tradeoff} visualizes these constraints for an example user.

\begin{figure}
\centering
\captionsetup[subfigure]{font=scriptsize,labelformat=empty,aboveskip=1pt,belowskip=2pt}
  \centering
  
  \begin{subfigure}[c]{\columnwidth}
    \centering\includegraphics[width=\textwidth]{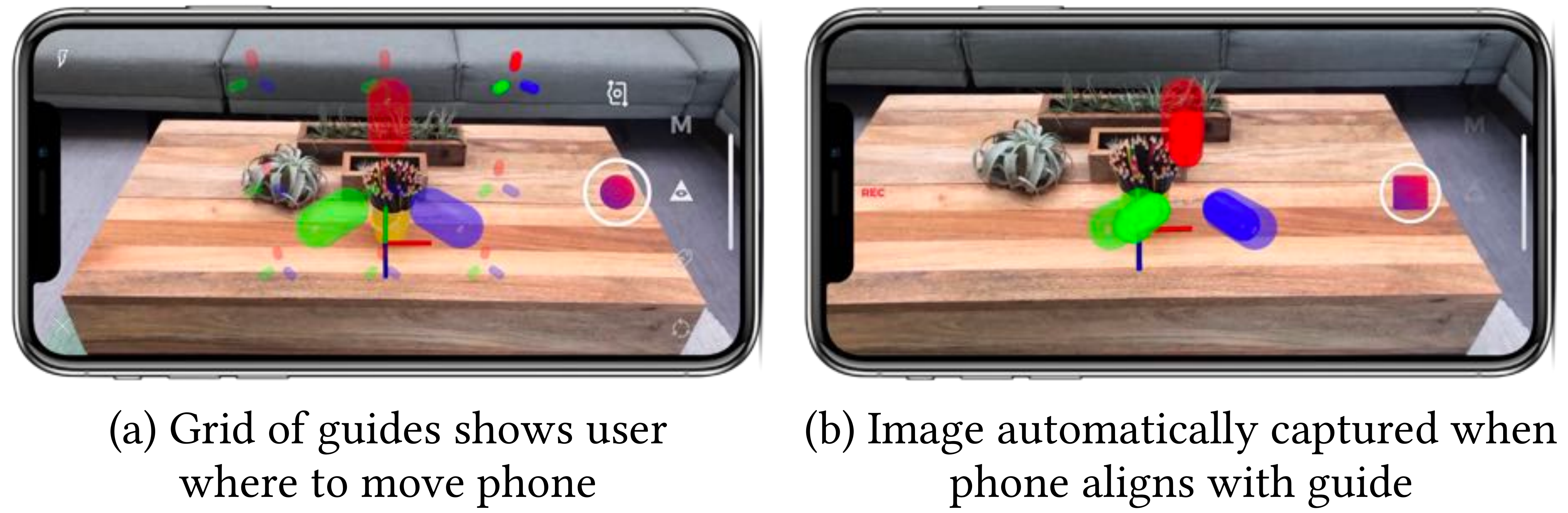}
  \end{subfigure}
    \caption{Equation~\ref{eqn:sampling_req} prescribes a simple sampling bound related only to the maximum scene disparity. We take advantage of the augmented reality toolkits available in modern smartphones to create an app that helps the user sample a real scene for rendering with our method. (a) We use built-in software to track the phone's position and orientation, providing sampling guides that allow the user to space photos evenly at the target disparity. (b)~Once the user has centered the phone so that the RGB axes align with one of the guides, the app automatically captures a photo.}
    \label{fig:phone_app}
\end{figure}

\subsection{Smartphone Capture App}
\label{sec:app}

We develop an app for iOS smartphones, based on the ARKit framework, that guides users to capture input views for our view synthesis algorithm. The user first taps the screen to mark the closest object, and the app uses the corresponding scene depth computed by ARKit as $z_{\min}$. Next, the user selects the size of the view plane $S$ within which our algorithm will render novel views. We fix the rendering resolution for the smartphone app to $W=500$ which therefore fixes the prescribed number and spacing of required images based on Equation~\ref{eqn:smartphone_guideline} and the definition $\Delta_u=S/\sqrt{N}$. Our app then guides the user to capture these views using the intuitive augmented reality overlay shown in Figure~\ref{fig:phone_app}. When the phone detects that the camera has been moved to a new sample location, it automatically records an image and highlights the next sampling point. 

\subsection{Preprocessing}
\label{sec:preprocess}

After capturing the required input images, the only preprocessing required before being able to render novel views is estimating the input camera poses and using our trained network to predict an MPI for each input view. Unfortunately, camera poses from ARKit are currently not accurate enough for acceptable results, so we use the open source COLMAP software package~\cite{schoenberger2016sfm, schoenberger2016mvs}, which takes about 2-6 minutes for sets of 20-30 input images. 

We use the deep learning pipeline described in Section~\ref{sec:cnn} to predict an MPI for each input sampled view. On an Nvidia GTX 1080Ti GPU, This takes approximately 0.5 seconds for a small MPI ($500\times 350 \times 32 \approx 6$ megavoxels) or 12 seconds for a larger MPI that must be output in overlapping patches ($1000\times 700 \times 64 \approx 45$ megavoxels). In total, our method only requires about 10 minutes of preprocessing to estimate poses and predict MPIs before being able to render novel views at a 1 megapixel image resolution.

With the increasing investment in smartphone AR and on-device deep learning accelerators, we expect that smartphone pose estimation will soon be accurate enough and on-device network inference will be powerful enough for users to go from capturing images to rendering novel views within a few seconds.

\subsection{Real-Time Viewers}
\label{sec:viewer}

We implement novel view rendering from a single MPI by rasterizing each plane from back to front using texture mapped rectangles in 3D space, invoking a standard shader API to correctly handle the alpha compositing, perspective projection, and texture resampling. For each new view, we determine the MPIs to be blended, as discussed in Section~\ref{sec:blending}, and render them into separate framebuffers. We then use a simple fragment shader to perform the alpha-weighted blending described in Section~\ref{sec:blending}. We implement this rendering pipeline as desktop viewer using OpenGL which renders views with $1000\times 700$ resolution at 60 frames per second, as well as an iOS mobile viewer using the Metal API which renders views with $500\times 350$ resolution at 30 frames per second. Please see our video for demonstrations of these real-time rendering implementations.

\subsection{Limitations}

A main limitation of our algorithm is that the MPI network sometimes assigns high opacity to incorrect layers in regions of ambiguous or repetitive texture and regions where scene content moves between input images. This can cause floating or blurred patches in the rendered output sequence (see the far right side of the fern in our video), which is a common failure mode in methods that rely on texture matching cues to infer depth. These artifacts could potentially be ameliorated by using more input views to disambiguate stereo matching and by encouraging the network to learn stronger global priors on 3D geometry.

Another limitation is the difficulty of scaling to higher image resolutions. As evident in Equations~\ref{eq:time_complexity} and~\ref{eq:space_complexity}, layered approaches such as our method are limited by complexities that scale cubically with the image width in pixels. Furthermore, increasing the image resolution requires a CNN with a larger receptive field. This could be addressed by exploring multiresolution CNN architectures and hierarchical volume representations such as octrees, or by predicting a more compact local scene representation such as layered depth images~\cite{shade98} with opacity. 

\section{Conclusion}

We have presented a simple and practical method for view synthesis that works reliably for complex real-world scenes, including non-Lambertian materials. Our algorithm first promotes each input image into a layered local light field representation, then renders novel views in real time by blending outputs generated by nearby representations. We extend traditional layered plenoptic sampling analysis to handle occlusions and provide a theoretical sampling bound on how many views are needed for our method to render high-fidelity views of a given scene. We quantitatively validate this bound and demonstrate that we match the perceptual quality of dense Nyquist rate view sampling while using $\approx4000\times$ fewer input images. Our accompanying video demonstrates that we thoroughly outperform prior work, and showcases results on over 60 diverse and complex real-world scenes, where our novel views are rendered with a fully automated capture-to-render pipeline. We believe that our work paves the way for future advances in image-based rendering that combine the empirical performance benefits of data-driven machine learning methods with the robust reliability guarantees of traditional geometric and signal processing based analysis.

\begin{acks}
We thank the SIGGRAPH reviewers for their constructive comments. The technical video was created with help from Julius Santiago, Milos Vlaski, Endre Ajandi, and Christopher Schnese. The augmented reality app was developed by Alex Trevor. 

Ben Mildenhall is funded by a Hertz Foundation Fellowship. Pratul P. Srinivasan is funded by an NSF Graduate Fellowship. Ravi Ramamoorthi is supported in part by NSF grant 1617234, ONR grant N000141712687, and Google Research Awards. Ren Ng is supported in part by NSF grant 1617794 and an Alfred P. Sloan Foundation Fellowship.
\end{acks}

\appendix

\begin{table}[t]
\centering{
\caption{Our network architecture. \textbf{k} is the kernel size, \textbf{s} the stride, \textbf{d} the kernel dilation, \textbf{chns} the number of input and output channels for each
layer, \textbf{in} and \textbf{out} are the accumulated stride for the input and output of each layer, \textbf{input} denotes the input of each layer with $+$ meaning concatenation, and layers starting with ``nnup'' perform $2\times$ nearest neighbor upsampling.}
\label{tab:net}
\resizebox{\columnwidth}{!}{
\begin{tabular}{cccccccc}
\toprule
  \textbf{Layer} & \textbf{k} & \textbf{s} & \textbf{d} & \textbf{chns} & \textbf{in} & \textbf{out} & \textbf{input}
\tabularnewline
\midrule
conv1\_1 & 3 & 1 & 1 & 15/8   & 1 & 1 & PSVs
\tabularnewline
conv1\_2 & 3 & 2 & 1 & 8/16  & 1 & 2 & conv1\_1
\tabularnewline
conv2\_1 & 3 & 1 & 1 & 16/16 & 2 & 2 & conv1\_2
\tabularnewline
conv2\_2 & 3 & 2 & 1 & 16/32 & 2 & 4 & conv2\_1
\tabularnewline
conv3\_1 & 3 & 1 & 1 & 32/32 & 4 & 4 & conv2\_2
\tabularnewline
conv3\_2 & 3 & 1 & 1 & 32/32 & 4 & 4 & conv3\_1
\tabularnewline
conv3\_3 & 3 & 2 & 1 & 32/64 & 4 & 8 & conv3\_2
\tabularnewline
conv4\_1 & 3 & 1 & 2 & 64/64 & 8 & 8 & conv3\_3
\tabularnewline
conv4\_2 & 3 & 1 & 2 & 64/64 & 8 & 8 & conv4\_1
\tabularnewline
conv4\_3 & 3 & 1 & 2 & 64/64 & 8 & 8 & conv4\_2
\tabularnewline
\midrule
nnup5 &  &  &  & 128/256 & 8 & 4 & conv4\_3 + conv3\_3
\tabularnewline
conv5\_1 & 3 &  1 & 1 & 256/32 & 4 & 4 & nnup5
\tabularnewline
conv5\_2 & 3 &  1 & 1 &  32/32 & 4 & 4 & conv5\_1
\tabularnewline
conv5\_3 & 3 &  1 & 1 &  32/32 & 4 & 4 & conv5\_2
\tabularnewline
nnup6 &  &  &  &  64/128 & 4 & 2 & conv5\_3 + conv2\_2
\tabularnewline
conv6\_1 & 3 &  1 & 1 &  128/16 & 2 & 2 & nnup6
\tabularnewline
conv6\_2 & 3 &  1 & 1 &  16/16 & 2 & 2 & conv6\_1
\tabularnewline
nnup7 &  &  &  &   32/64 & 2 & 1 & conv6\_2 + conv1\_2
\tabularnewline
conv7\_1 & 3 &  1 & 1 &   64/8 & 1 & 1 & nnup7
\tabularnewline
conv7\_2 & 3 &  1 & 1 &   8/8  & 1 & 1 & conv7\_1
\tabularnewline
conv7\_3 & 3 &  1 & 1 &   8/5  & 1 & 1 & conv7\_2
\tabularnewline
\bottomrule
\end{tabular}}
}
\end{table}

            




\section{Baseline Methods Implementation Details}
\label{sec:baseline_implementation}

Here we provide additional implementation details regarding the baseline comparison methods described in Section~\ref{sec:baselines}.

\paragraph{Soft3D~\cite{penner17}} We implemented this algorithm from the description provided in the original paper, since no open-source code is currently available. We provide Soft3D's local stereo stage with the same 5 input images used to compute each of our MPIs, for fair comparison. Next, we aggregate Soft3D's computed vote volumes across 25 viewpoints, as suggested in their paper for 2D view captures. Finally, we compute each rendered novel view using the Soft3D volumes corresponding to the same 5 viewpoints whose MPIs we blend between for our algorithm. The guided filter parameters are not specified in the original Soft3D paper, but we find that best results were obtained using a window size of 8 and $\epsilon=20$ for images with values in the range $(0,255)$. Please refer to our supplemental video to see results of our Soft3D implementation on a scene shown in the original paper's video to verify that our implementation is of similar quality. 

\paragraph{Backwards warping deep network (BW Deep)} This baseline is very similar to~\cite{kalantari16} in overall structure. However, it uses a much larger CNN architecture (same architecture as our method) and takes five images as input rather than four. In addition, since our network is 3D rather than 2D, we can run inference with a variable number of planes in the plane sweep volumes rather than fixing $D=100$. When generating our quantitative results (Table~\ref{table:synth_quant}), we set $D=d_{\max}$, as in our method. We train this baseline with the same synthetic and real datasets as our network.

\paragraph{Unstructured Lumigraph Rendering (ULR)~\cite{gortler96, buehler01}} We use COLMAP's multiview stereo implementation~\cite{schoenberger2016sfm,schoenberger2016mvs} to generate dense depths and estimate a global triangle mesh of the scene from all input images using the screened Poisson surface reconstruction algorithm~\cite{kazhdan13}. For each pixel in a new target view, we implement the heuristic blending weights from~\cite{buehler01} to blend input images reprojected using the global mesh geometry. 

\paragraph{Light Field Interpolation (LFI)~\cite{chai00}} This baseline is representative of classic signal processing based continuous view reconstruction. Following the method of plenoptic sampling~\cite{chai00}, for data on a regular 2D grid, we render novel views using a bilinear interpolation reconstruction filter sheared to the mean scene disparity. For unstructured real data, we blend together 5 nearby views reprojected to the mean scene disparity, using the same blending weights as our own method.

Note that for methods that depend on a plane sweep volume (Soft3D, BW Deep, and all versions of our method), we set the number of depth planes used in the PSVs to $d_{\max}$.

\section{Network Architecture}
\label{sec:network_architecture}

Table~\ref{tab:net} contains a detailed specification of our 3D CNN architecture. All layers except the last are followed by a ReLU nonlinearity and layer normalization~\cite{layernorm}. The final layer outputs 5 channels. One channel is passed through a sigmoid to generate the output MPI's alpha channel. The other four (along with an all-zero channel) are passed through a softmax to get five blending weights for each voxel which are used to generate the output MPI's color channels, as described in Section~\ref{sec:cnn}. Restricting one of the softmax inputs to always be zero makes the function one-to-one rather than many-to-one.

\bibliographystyle{ACM-Reference-Format}
\bibliography{main}

\end{document}